%% file: arxiv.tex
\documentclass[10pt,journal,compsoc]{IEEEtran}
\ifCLASSOPTIONcompsoc
  \usepackage[nocompress]{cite}
\else
  \usepackage{cite}
\fi

\ifCLASSINFOpdf
\else
\fi

\usepackage{url}
\usepackage{graphicx}
\usepackage{subfigure}
\usepackage{threeparttable}
\usepackage{amsmath}
\usepackage{color}
\usepackage{xcolor}
\usepackage{wrapfig}
\usepackage{booktabs}
\usepackage{multirow}
\usepackage{capt-of}
\usepackage{enumerate}
\usepackage{tabu}
\usepackage{array}
\usepackage[ruled]{algorithm2e}
\input{math_commands.tex}

\makeatletter
\DeclareRobustCommand\onedot{\futurelet\@let@token\@onedot}
\def\@onedot{\ifx\@let@token.\else.\null\fi\xspace}
\def\eg{\emph{e.g}\onedot} 
\def\ie{\emph{i.e}\onedot} 
 
\def\etc{\emph{etc}\onedot}

\newcommand*{\@rowstyle}{}
\newcommand*{\rowstyle}[1]{% sets the style of the next row
  \gdef\@rowstyle{#1}%
  \@rowstyle\ignorespaces%
}
\newcolumntype{=}{% resets the row style
  >{\gdef\@rowstyle{}}%
}
\newcolumntype{+}{% adds the current row style to the next column
  >{\@rowstyle}%
}
\makeatother

\def\usecolor{1}
\if\usecolor1
    \definecolor{revcolor}{rgb}{0,0,0}
    \usepackage[colorlinks=true]{hyperref}
\else
    \definecolor{revcolor}{rgb}{0,0,0}
    \usepackage[colorlinks=false]{hyperref}
    \hypersetup{hidelinks}
\fi

\newcommand{\name}{FNA++}
\newcommand{\rev}[1]{\textcolor{revcolor}{#1}}

\begin{document}

\title{\name: Fast Network Adaptation via Parameter Remapping and Architecture Search}

\author{
    Jiemin Fang$^*$, Yuzhu Sun$^*$, Qian Zhang, Kangjian Peng, Yuan Li, Wenyu Liu,~\IEEEmembership{Senior Member,~IEEE}, and Xinggang Wang$^\dagger$,~\IEEEmembership{Member,~IEEE}
    \IEEEcompsocitemizethanks{
        \IEEEcompsocthanksitem J. Fang is with the Institute of Artificial Intelligence and the School of Electronic Information and Communications, Huazhong University of Science and Technology, Wuhan 430074, P.R. China.
        \IEEEcompsocthanksitem Y. Sun, W. Liu and X. Wang are with the School of Electronic Information and Communications, Huazhong University of Science and Technology, Wuhan 430074, P.R. China. The work was done during the internship of J. Fang and Y. Sun at Horizon Robotics.
        \IEEEcompsocthanksitem  Q. Zhang and K. Peng are with Horizon Robotics.
        \IEEEcompsocthanksitem Y. Li is with Google.
    }
    \thanks{$^*$ J. Fang and Y. Sun  contribute equally to this work.}
    \thanks{$^\dagger$ Corresponding to X. Wang (\url{xgwang@hust.edu.cn}).}
}

% \markboth{Journal of \LaTeX\ Class Files,~Vol.~14, No.~8, August~2015}%
% {Shell \MakeLowercase{\textit{et al.}}: Bare Demo of IEEEtran.cls for Computer Society Journals}

\IEEEtitleabstractindextext{%
\begin{abstract}
Deep neural networks achieve remarkable performance in many computer vision tasks. Most state-of-the-art (\emph{SOTA}) semantic segmentation and object detection approaches reuse neural network architectures designed for image classification as the backbone, commonly pre-trained on ImageNet. However, performance gains can be achieved by designing network architectures specifically for detection and segmentation, as shown by recent neural architecture search (NAS) research for detection and segmentation. One major challenge though is that ImageNet pre-training of the search space representation (a.k.a.~super network) or the searched networks incurs huge computational cost. In this paper, we propose a Fast Network Adaptation (\name) method, which can adapt both the architecture and parameters of a seed network (e.g. an ImageNet pre-trained network) to become a network with different depths, widths, or kernel sizes via a \emph{parameter remapping} technique, making it possible to use NAS for segmentation and detection tasks a lot more efficiently. In our experiments, we apply \name\ on MobileNetV2 to obtain new networks for semantic segmentation, object detection, and human pose estimation that clearly outperform existing networks designed both manually and by NAS. We also implement \name\ on ResNets and NAS networks, which demonstrates a great generalization ability. The total computation cost of \name\ is significantly less than \emph{SOTA} segmentation and detection NAS approaches: 1737$\times$ less than DPC, 6.8$\times$ less than Auto-DeepLab, and 8.0$\times$ less than DetNAS. A series of ablation studies are performed to demonstrate the effectiveness, and detailed analysis is provided for more insights into the working mechanism. Codes are available at \url{https://github.com/JaminFong/FNA}.
\end{abstract}

\begin{IEEEkeywords}
Fast network adaptation, parameter remapping, neural architecture search.
\end{IEEEkeywords}}

\maketitle

\IEEEdisplaynontitleabstractindextext

% \fi
\IEEEpeerreviewmaketitle

\IEEEraisesectionheading{\section{Introduction}\label{sec:introduction}}

\IEEEPARstart{D}{eep} convolutional neural networks have achieved great successes in computer vision tasks such as image classification~\cite{krizhevsky2012imagenet, he2016deep, howard2017mobilenets}, semantic segmentation~\cite{long2015fully, ronneberger2015u, deeplab-v3}, object detection~\cite{ren2015faster, liu2016ssd, lin2017focal} and pose estimation~\cite{xiao2018simple, sun2019deep} etc. Image classification has always served as a fundamental task for neural architecture design. It is common to use networks designed and pre-trained on the classification task as the backbone and fine-tune them for segmentation or detection tasks. However, the backbone plays an important role in the performance on these tasks and the difference between these tasks calls for different design principles of the backbones. For example, segmentation tasks require high-resolution features and object detection tasks need to make both localization and classification predictions from each convolutional feature. Such distinctions make neural architectures designed for classification tasks fall short. Some attempts~\cite{DBLP:conf/eccv/LiPYZDS18, DBLP:journals/corr/abs-1908-07919} have been made to tackle this problem by manually modifying the architectures designed for classification to better accommodate to the characteristics of new tasks.

Handcrafted neural architecture design is inefficient, requires a lot of human expertise, and may not find the best-performing networks. Recently, neural architecture search (NAS) methods~\cite{zoph2017learning, pham2018efficient, liu2017progressive} see a rise in popularity. Automatic deep learning methods aim at helping engineers get rid of tremendous trial and error on architecture designing and further promoting the performance of architectures over manually designed ones. Early NAS works~\cite{zoph2017learning, MnasNet,cai2018proxylessnas} explore the search problem on the classification tasks. As the NAS methods develop, some works~\cite{liu2019auto, zhang2019customizable, DBLP:journals/corr/abs-1903-10979} propose to use NAS to specialize the backbone architecture design for semantic segmentation or object detection tasks. Nevertheless, backbone pre-training remains an inevitable but costly procedure. Though some works like~\cite{DBLP:journals/corr/abs-1811-08883} recently demonstrate that pre-training is not always necessary for accuracy considerations, training from scratch on the target task still takes more iterations than fine-tuning from a pre-trained model. For NAS methods, the pre-training cost is non-negligible for evaluating the networks in the search space. One-shot search methods~\cite{brock2017smash, Understanding, DBLP:journals/corr/abs-1903-10979} integrate all possible architectures in one super network but pre-training the super network and the searched network still bears huge computation cost.

\begin{figure*}[tbp]
    \begin{center}
        \includegraphics[width=1.0\linewidth]{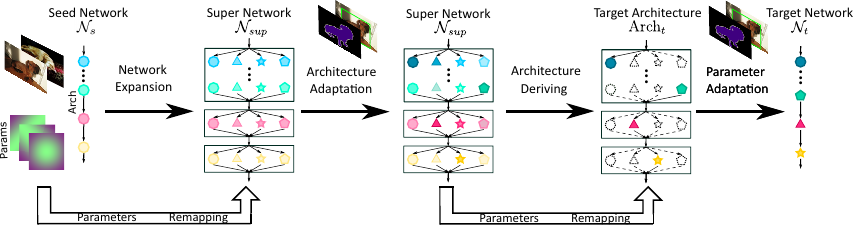}
    \end{center}
    \caption{The framework of our proposed \name. Firstly, we select an pre-trained network as the seed network $\gN_s$ and expand $\gN_s$ to a super network $\gN_{sup}$ which is the representation of the search space. The parameters of $\gN_s$ are remapped to $\gN_{sup}$ and architecture adaptation is performed. Then we derive the target architecture $\mathit{Arch}_t$ based on the parameter distribution in $\gN_{sup}$. Before parameter adaptation, we remap the parameters of $\gN_{sup}$ to $\mathit{Arch}_t$. Finally, we adapt the parameters of $\mathit{Arch}_t$ to get the target network $\gN_t$.}
    \label{fig: framework}
\end{figure*}

As ImageNet~\cite{imagenet} pre-training has been a standard practice for many computer vision tasks, there are lots of models trained on ImageNet available in the community. To take full advantages of these pre-trained models, we propose a Fast Network Adaptation (\name) method based on a novel parameter remapping paradigm. Our method can adapt both the architecture and parameters of one network to a new task with negligible cost. Fig.~\ref{fig: framework} shows the whole framework. The adaptation is performed on both the architecture- and parameter-level. We adopt the NAS methods~\cite{zoph2017learning, Real2018Regularized, liu2018darts} to implement the architecture-level adaptation. We select the manually designed network as the seed network, which is pre-trained on ImageNet. Then, we expand the seed network to a super network which is the representation of the search space in \name. We initialize new parameters in the super network by mapping those from the seed network using the proposed parameter remapping mechanism. Compared with previous NAS methods~\cite{DBLP:conf/nips/ChenCZPZSAS18,liu2019auto,DBLP:journals/corr/abs-1903-10979} for segmentation or detection tasks that search from scratch, our architecture adaptation is much more efficient thanks to the parameter remapped super network. With architecture adaptation finished, we obtain a target architecture for the new task. Similarly, we remap the parameters of the super network which are trained during architecture adaptation to the target architecture. Then we fine-tune the parameters of the target architecture on the target task with no need of backbone pre-training on a large-scale dataset.

We demonstrate \name's effectiveness and efficiency via experiments on semantic segmentation, object detection and human pose estimation tasks. We adapt the manually designed network MobileNetV2~\cite{sandler2018mobilenetv2} to the semantic segmentation framework DeepLabv3~\cite{deeplab-v3}, object detection framework RetinaNet~\cite{lin2017focal} and SSDLite~\cite{liu2016ssd, sandler2018mobilenetv2} and human pose estimation framework SimpleBaseline~\cite{xiao2018simple}. Networks adapted by \name\ surpass both manually designed and NAS networks in terms of both performance and model MAdds. Compared to NAS methods, \name\ costs 1737$\times$ less than DPC~\cite{DBLP:conf/nips/ChenCZPZSAS18}, 6.8$\times$ less than Auto-DeepLab~\cite{liu2019auto} and 8.0$\times$ less than DetNAS~\cite{DBLP:journals/corr/abs-1903-10979}. To demonstrate the generalizability of our method, we implement \name\ on diverse networks, including ResNets~\cite{he2016deep} and NAS networks, \ie, FBNet~\cite{fbnet} and ProxylessNAS~\cite{cai2018proxylessnas}, which are searched on the ImageNet classification task. Experimental results show that \name\ can further promote the performance of ResNets and NAS networks on the new task (object detection in our experiment).

Our main contributions can be summarized as follows:
\begin{itemize}
    \item We propose a novel \name\ method that automatically fine-tunes both the architecture and the parameters of an ImageNet pre-trained network on target tasks. \name\ is based on a novel parameter remapping mechanism which is performed for both architecture adaptation and parameter adaptation. 
    \item \name\ promotes the performance on semantic segmentation, object detection and human pose estimation tasks with much lower computation cost than previous NAS methods, \eg 1737$\times$ less than DPC, 6.8$\times$ less than Auto-DeepLab and 8.0$\times$ less than DetNAS.
    % outperforms the previous manual\footnote{\xw{there are many manually designed networks; we cannot say that we outperform all manually designed networks.}} and NAS-based semantic segmentation, object detection and human pose estimation methods with much lower computation costs, \eg 1737$\times$ less than DPC, 6.8$\times$ less than Auto-DeepLab and 8.0$\times$ less than DetNAS. 
    \item \name\ is generally helpful for various visual recognition tasks and improves over various pre-trained networks, \eg, MobileNets, ResNets and NAS networks (FBNet~\cite{fbnet} and ProxylessNAS~\cite{cai2018proxylessnas}).
\end{itemize}

Our preliminary version of this manuscript was previously published as a conference paper~\cite{Fang*2020Fast}. We make improvements to the former version as follows. First, we generalize the paradigm of parameter remapping and now it is applicable to more architectures, \eg, ResNet~\cite{he2016deep} and NAS networks with various depths, widths and kernel sizes. Second, we improve the remapping mechanism for parameter adaptation and achieve better results than our former version over different frameworks and tasks with no computation cost increased. Third, we implement \name\ on one more task (SimpleBaseline for human pose estimation) and achieve great performance. \rev{Fourth, we provide more theoretical analysis and discussions about the proposed method, which reveal more on the working mechanism of our method. Finally, we provide more comprehensive studies on our method, including detailed analysis on the searched architectures, theoretical and empirical comparisons between our proposed parameter remapping and Net2Net, super network pre-training and additional random remapping experiments, multiple runs for reporting more stable results \etc.}
% The capacity of handling more networks and broader range of applications with stronger performance make our method more professional; thus we name our method in this version as \name.

The remaining part of the paper is organized as follows. In Sec.~\ref{Sec: RelatedWork}, we describe the related works from three aspects, neural architecture search, backbone design and parameter remapping. Then we introduce our method in Sec.~\ref{sec: Method}, including the proposed parameter remapping mechanism and the detailed adaptation process. In Sec.~\ref{sec: Experiments}, we evaluate our method on different tasks and frameworks. The method is also implemented on various networks. \rev{Then in Sec.~\ref{sec: ablat}, a series of experiments are performed to study the proposed method comprehensively. We further provide more theoretical analysis and discussion about the working mechanism in Sec.~\ref{sec: work_mech}.} We finally conclude in Sec.~\ref{sec: Conclusion}.

% --------------------------------Section Partition------------------------------- %
\section{Related Work}
\label{Sec: RelatedWork}
    \subsection{Neural Architecture Search}
    Early NAS works automate network architecture design by applying the reinforcement learning (RL)~\cite{zoph2016neural, zoph2017learning, MnasNet} or evolutionary algorithm (EA)~\cite{liu2017hierarchical, Real2018Regularized} to the search process. The RL/EA-based methods obtain architectures with better performance than handcrafted ones but usually bear tremendous search cost. Afterwards, ENAS~\cite{pham2018efficient} proposes to use parameter sharing to decrease the search cost but the sharing strategy may introduce inaccuracy on evaluating the architectures. NAS methods based on the one-shot model~\cite{brock2017smash, Understanding, dong2019one} lighten the search procedure by introducing a super network as a representation of all possible architectures in the search space. Recently, differentiable NAS~\cite{liu2018darts,cai2018proxylessnas,fbnet,fang2020densely,dong2019searching} arises great attention in this field which achieves remarkable results with far lower search cost compared with previous ones. Differentiable NAS assigns architecture parameters to the super network and updates the architecture parameters by gradient descent. The final architecture is derived based on the distribution of architecture parameters. We use the differentiable NAS method to implement network architecture adaptation, which adjusts the backbone architecture automatically to new tasks with remapped seed parameters accelerating. In experiments, we perform random search and still achieve great performance, which demonstrates \name\ is agnostic of NAS methods and can be equipped with diverse NAS methods.

    \subsection{Backbone Design}
    As deep neural network designing~\cite{simonyan2014very, szegedy2016rethinking, he2016deep} develops, the backbones of semantic segmentation or object detection networks evolve accordingly. Most previous methods~\cite{ren2015faster,lin2017focal,liu2016ssd,deeplab-v3} directly reuse the networks designed on classification tasks as the backbones. However, the reused architecture may not meet the demands of the new task characteristics. Some works improve the backbone architectures by modifying existing networks. PeleeNet~\cite{wang2018pelee} proposes a variant of DenseNet~\cite{huang2017densely} for more real-time object detection on mobile devices. DetNet~\cite{DBLP:conf/eccv/LiPYZDS18} applies dilated convolutions~\cite{yu2015multi} in the backbone to enlarge the receptive field which helps to detect objects more precisely. BiSeNet~\cite{yu2018bisenet} and HRNet~\cite{DBLP:journals/corr/abs-1908-07919} design multiple paths to learn both high- and low- resolution representations for better dense prediction. Recently, some works propose to use NAS methods to redesign the backbone networks automatically. Auto-DeepLab~\cite{liu2019auto} searches for architectures with cell structures of diverse spatial resolutions under a hierarchical search space. The searched resolution change patterns benefit to dense image prediction problems. CAS~\cite{zhang2019customizable} proposes to search for the semantic segmentation architecture under a lightweight framework while the inference speed optimization is considered. DetNAS~\cite{DBLP:journals/corr/abs-1903-10979} searches for the backbone of the object detection network under a ShuffleNet~\cite{zhang2017shufflenet,DBLP:journals/corr/abs-1807-11164}-based search space. They use the one-shot NAS method to decrease the search cost. However, pre-training the super network on ImageNet and the final searched network bears a huge cost. Benefiting from the proposed parameter remapping mechanism, our \name\ adapts the architecture to new tasks with a negligible cost.

    \subsection{Parameter Remapping}
    Net2Net~\cite{chen2015net2net} proposes the function-preserving transformations to remap the parameters of one network to a new deeper or wider network. This remapping mechanism accelerates the training of the new larger network and achieves great performance. \rev{Following this manner, EAS~\cite{cai2018efficient} uses the function-preserving transformation to grow the network depth or layer width, and TreeCell~\cite{cai2018path} performs a path-level transformation on the tree structure for architecture search.} The computation cost can be saved by reusing the weights of previously validated networks. Moreover, some NAS works~\cite{pham2018efficient, eatnas, elsken2018efficient} apply parameter sharing on child models to accelerate the search process while the sharing strategy is intrinsically parameter remapping. \rev{Some methods, \eg, \cite{stamoulis2019single,chen2019efficient}, share the parameters on the kernel level in a one-shot model. Once-for-All~\cite{cai2019once} transforms the parameters of the super network to sub-networks and obtain various target networks without additional training.} Our parameter remapping paradigm extends the mapping dimension to the depth-, width- and kernel- level. Compared to Net2Net which only focuses on mapping parameters to a deeper and wider network, the remapping mechanism in \name\ has more flexibility and can be performed on architectures with various depths, widths and kernel sizes. The remapping mechanism transfers the information between different tasks and helps both the architecture and parameter adaptation achieve great performance with low computation cost.

% --------------------------------Section Partition------------------------------- %

\begin{figure*}[thbp]
    \centering
    \subfigure[Depth level]
    {\begin{minipage}[b]{0.2\textwidth}
        \includegraphics[width=\textwidth]{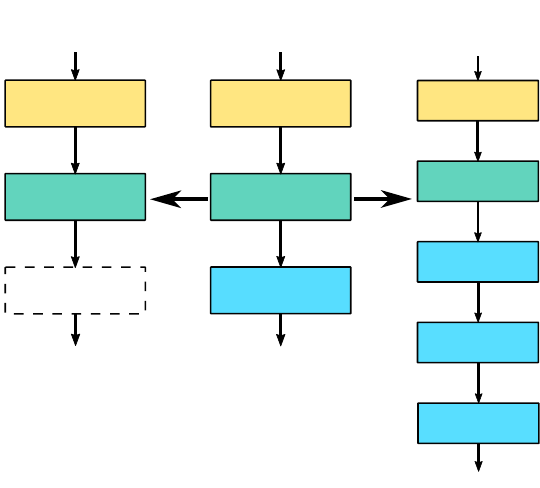}
        \label{fig: depth_mapping}
    \end{minipage}}
    \hfill
    \subfigure[Width level]
    {\begin{minipage}[b]{0.4\textwidth}
        \includegraphics[width=\textwidth]{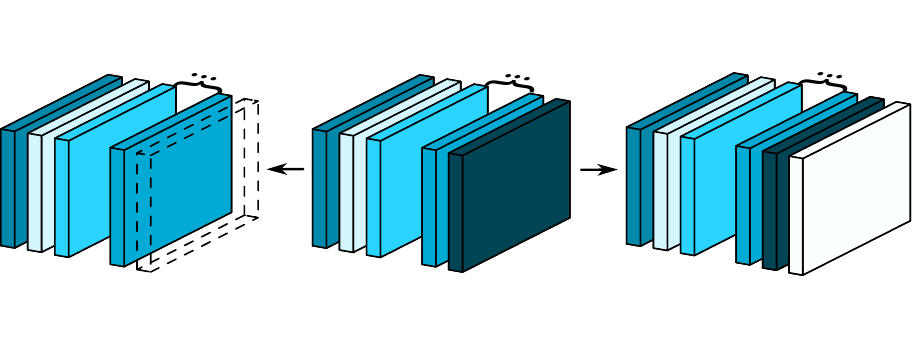}
        \label{fig: channel_mapping}
    \end{minipage}}
    \hfill
    \subfigure[Kernel level]
    {\begin{minipage}[b]{0.28\textwidth}
        \includegraphics[width=\textwidth]{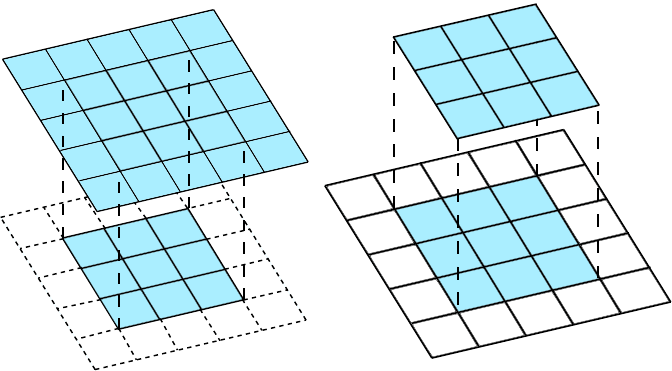}
        \label{fig: kernel_mapping}
    \end{minipage}}
    \caption{Parameters are remapped on three levels. (a) shows the depth-level remapping. The parameters of the new network are remapped from the corresponding layers in the seed network. The parameters of new layers are remapped from the last layer in the seed network.
    (b) shows the width-level remapping. For parameters with fewer channels (\emph{left}), the seed parameters are remapped to the new network with corresponding channels. For parameters with more channels (\emph{right}), the seed parameters are remapped to corresponding channels and parameters in new channels are assigned with $0$ (denoted as the white cuboid).
    (c) shows the kernel-level remapping. For parameters in a smaller kernel (left), the central part of seed parameters are remapped to the new kernel. For parameters in a larger kernel (right), the seed parameters are remapped to the central part of the new kernel. The values of the other part are assigned with $0$.}
\end{figure*}

\section{Method}
\label{sec: Method}
In this section, we first introduce the proposed parameter remapping paradigm, which is performed on three levels, \ie, network depth, layer width and convolution kernel size. Then we explain the whole procedure of the network adaptation including three main steps, network expansion, architecture adaptation and parameter adaptation. The parameter remapping paradigm is applied before architecture and parameter adaptation.

% \begin{figure}[htbp]
%     \centering
%         \includegraphics[width=0.7\columnwidth]{figs/mbconv.pdf}
%         \caption{Inverted residual block (MBConv)~\cite{sandler2018mobilenetv2}. The normal block (\emph{upper}) inputs and outputs the tensor with the same channel number and spatial resolution. The residual connection is used in the normal block. The reduction block (\emph{bottom}) performs the down-sampling operation with stride 2 and transforms the channel number.}
% \label{fig: mbconv}
% \end{figure}
    
\subsection{Parameter Remapping}
\label{subsec: params_remap}
We define \emph{parameter remapping} as one paradigm which maps the parameters of one seed network to another one. We denote the seed network as $\gN_s$ and the new network as $\gN_n$, whose parameters are denoted as $\mW_s$ and $\mW_n$ respectively. The remapping paradigm is illustrated in the following three aspects. The remapping on the depth-level is firstly carried out and then the remapping on the width- and kernel- level is conducted simultaneously. Moreover, we study different remapping strategies in the experiments (Sec.~\ref{subsec: studyPR}).

    \subsubsection{Remapping on Depth-level}
    We introduce diverse depth settings in our architecture adaptation process. Specifically, we adjust the number of MobileNetV2~\cite{sandler2018mobilenetv2} or ResNet~\cite{he2016deep} blocks in every stage of the network. We assume that one stage in the seed network $\gN_s$ has $l$ layers. The parameters of each layer can be denoted as $\{\mW_s^{(1)}, \mW_s^{(2)}, \dots, \mW_s^{(l)}\}$. Similarly, we assume that the corresponding stage with $m$ layers in the new network $\gN_n$ has parameters $\{\mW_n^{(1)}, \mW_n^{(2)}, \dots, \mW_n^{(m)}\}$. The remapping process on the depth-level is shown in Fig.~\ref{fig: depth_mapping}. The parameters of layers in $\gN_n$ which also exit in $\gN_s$ are just copied from $\gN_s$. The parameters of new layers are all copied from the last layer in the stage of $\gN_s$. Parameter remapping in layer $i$ is formulated as
    \begin{equation}
        \begin{aligned}
            f(i) &= min(i, l),  \\
            W_n^{(i)} &= W_s^{(f(i))},\quad \forall 1 \le i \le m \text{.}
        \end{aligned}
        \label{eq: depth_mapping}
    \end{equation}

    \subsubsection{Remapping on Width-level}
    In the MobileNetV2~\cite{sandler2018mobilenetv2} block, namely MBConv, the first point-wise convolution expands the low-dimensional features to a high dimension. This practice can be used for expanding the width and capacity of one neural network. We allow diverse expansion ratios for architecture adaptation. We denote the parameters of one convolution in $\gN_s$ as $\mW^s_{(out,in,h,w)} \in \R^{p \times q \times h \times w}$ and that in $\gN_n$ as $\mW^n_{(out,in,h,w)} \in \R^{r \times s \times h' \times w'}$, where $out$, $in$ denotes the output, input dimension of the parameter and $h, w$ denote the spatial dimension. The width-level remapping is illustrated in Fig.~\ref{fig: channel_mapping}. If the channel number of $\mW^n$ is smaller, the first $r$ or $s$ channels of $\mW^s$ are directly remapped to $\mW^n$. If the channel number of $\mW^n$ is larger than $\mW^s$, the parameters of $\mW^s$ are remapped to the first $p$ or $q$ channels in $\mW^n$. The parameters of the other channels in $\mW^n$ are initialized with $0$. The above remapping process can be formulated as follows.
    \begin{enumerate}[i]
        \item $r \leq p$:
        \begin{equation}
            \mW^n_{i,j,:,:} = \mW^s_{i,j,:,:}, \quad \forall 1 \le i \le r, 1 \le j \le s\text{,}
        \end{equation}
        \item $r > p$:
        \begin{equation}
            \mW^{n}_{i,j,:,:} = \begin{cases}
                \mW^{s}_{i,j,:,:} & \mbox{if } i \leq p \mbox{ and } j \leq q\\
                0 & \mbox{otherwise}
            \end{cases}\text{.}
        \end{equation}
    \end{enumerate}
    
    In our ResNet~\cite{he2016deep} adaptation, we allow architectures with larger receptive field by introducing grouped convolutions with larger kernel sizes, which do not introduce much additional MAdds. For architecture adaptation, the parameters $\mW^s \in \R^{p \times q \times h \times w}$ of the plain convolution in the seed network $\gN_s$ need to be remapped to the new parameters $\mW^n \in \R^{r \times s \times h \times w}$ of the grouped convolution in the super network $\gN_{sup}$. We assume the group number in the grouped convolution is $g$. The input channel number of the grouped convolution is $1/g$ of the plain convolution, \ie, $p = r$ and $s = q / g$. The parameters of the plain convolution $\mW^s$ are remapped to $\mW^n$ of the grouped convolution with the corresponding input dimension. This process can be formulated as, 
    \begin{equation}
        \begin{aligned}
            id_g &= \lfloor i / (p / g) \rfloor,\\
            \mW^n_{i,j,:,:} &= \mW^s_{i,id_g \times s + j,:,:}, \quad \forall 1 \le i \le r, 1 \le j \le s.
        \end{aligned}
    \end{equation}

    \subsubsection{Remapping on Kernel-level}
    The kernel size is commonly set as $3\times3$ in most artificially-designed networks~\cite{he2016deep,sandler2018mobilenetv2}. 
    However, the optimal kernel size settings may not be restricted to a fixed one. In a neural network, the larger kernel size can be used to expand the receptive field and capture abundant contextual features in segmentation or detection tasks but takes more computation cost than the smaller one. How to allocate the kernel sizes in a network more flexibly is explored in our method. We introduce the parameter remapping on the kernel size level and show it in Fig.~\ref{fig: kernel_mapping}. 
    We denote the weights of the convolution in the seed network $\gN_s$ as $\mW^{s, k}_{(out,in,h,w)}$ whose kernel size is $k \times k$. The weights in $\gN_n$ is denoted as $\mW^{n, k'}_{(out,in,h,w)}$ with $k' \times k'$ kernel size. 
    If the kernel size of $\mW^{n, k'}$ is smaller than $\mW^{s, k}$, the parameters of $\mW^{n, k'}$ are remapped from the central $k' \times k'$ region in $\mW^{s, k}$. Otherwise, we assign the parameters of the central $k \times k$ region in $\mW^{n, k'}$ with the values of $\mW^{s, k}$. The values of the other region surrounding the central part are assigned with $0$. The remapping process on the kernel-level is formulated as follows.\\
    \mbox{Let} $l = \frac{|k-k'|}{2}$,
    \begin{enumerate}[i]
        \item $k' \leq k$:
        \begin{equation}
            \mW^{n, k'}_{:,:,h,w} = \mW^{s, k}_{:,:,h+l,w+l}, \quad \forall 1 \leq h, w \leq k'\text{,}
        \end{equation}
        \item $k' > k$:
        \begin{equation}
            \mW^{n, k'}_{:,:,h,w} = \begin{cases}
                \mW^{s, k}_{:,:,h-l,w-l} & \mbox{if } l < h, w \le l+k\\
                0 & \text{otherwise}
            \end{cases}\text{,}
        \end{equation}
    \end{enumerate}
    where $h, w$ denote the indices of the spatial dimension.
    
\subsection{Fast Network Adaptation}
We divide our neural network adaptation into three steps. Fig.~\ref{fig: framework} illustrates the whole adaptation procedure. Firstly, we expand the seed network $\gN_s$ to a super network $\gN_{sup}$ which is the representation of the search space in the latter architecture adaptation process. Secondly, we perform the differentiable NAS method to implement network adaptation on the architecture-level and obtain the target architecture $\mathit{Arch}_t$. Finally, we adapt the parameters of the target architecture and obtain the target network $\gN_t$. The aforementioned parameter remapping mechanism is deployed before the two stages, \ie, architecture adaptation and parameter adaptation.

\begin{table}[t]
    \centering
    \caption{Search space with MobileNetV2~\cite{sandler2018mobilenetv2} as the seed network. ``chs": the number of output channels. ``n": the number of layers. ``s": the stride of the convolution. ``seg", ``det" and ``pose" denote the tasks of semantic segmentation, object detection and human pose estimation respectively. ``SBlock" denotes the block for search. }
    \begin{tabular}{c|c|c|c|c|c|c|c}
    \hline
    \multirow{2}*{\textbf{Block}} & \multirow{2}*{\textbf{chs}} & \multicolumn{3}{c|}{\textbf{n}} & \multicolumn{3}{c}{\textbf{s}} \\
    \cline{3-8}
    && seg & det & pose & seg & det & pose\\
    \hline
    $3 \times 3$ Conv & 32   & 1 & 1 & 1 & 2 & 2 & 2 \\
    MBConv(k3e1)      & 16   & 1 & 1 & 1 & 1 & 1 & 1 \\
    SBlock            & 24   & 4 & 4 & 4 & 2 & 2 & 2 \\
    SBlock            & 32   & 4 & 4 & 4 & 2 & 2 & 2 \\
    SBlock            & 64   & 6 & 4 & 4 & 2 & 2 & 2 \\
    SBlock            & 96   & 6 & 4 & 4 & 1 & 1 & 1 \\
    SBlock            & 160  & 4 & 4 & 4 & 1 & 2 & 2 \\
    SBlock            & 320  & 1 & 1 & 1 & 1 & 1 & 1 \\
    \hline    
    \end{tabular}
    \label{tab: space_mbv2}
\end{table}

    \subsubsection{Network Expansion}
    We expand the seed network $\gN_s$ to a super network $\gN_{sup}$ by introducing more options of architecture elements. For every MBConv layer, we allow for more kernel size settings $\{3, 5, 7\}$ and more expansion ratios $\{3, 6\}$. As most differentiable NAS methods~\cite{liu2018darts, cai2018proxylessnas, fbnet} do, we construct a super network as the representation of the search space. In the super network, we relax every layer by assigning each candidate operation with an architecture parameter. The output of each layer is computed as a weighted sum of output tensors from all candidate operations.
    \begin{equation}
        \overline{o}^{(i)}(x) = \sum_{o \in O} \frac{exp(\alpha^{(l)}_{o})}{\sum_{o' \in O} exp(\alpha^{(l)}_{o'})}o(x)\text{,}
    \end{equation}
    where $O$ denotes the operation set, $\alpha_o^{(l)}$ denotes the architecture parameter of operation $o$ in the $l$th layer, and $x$ denotes the input tensor. We set more layers in one stage of the super network and add the identity connection to the candidate operations for depth search. The structure of the search space is detailed in Tab.~\ref{tab: space_mbv2}. After expanding the seed network $\gN_s$ to the super network $\gN_{sup}$, we remap the parameters of $\gN_s$ to $\gN_{sup}$ based on the paradigm illustrated in Sec.~\ref{subsec: params_remap}. As shown in Fig.~\ref{fig: framework}, the parameters of different candidate operations (except the identity connection) in one layer of $\gN_{sup}$ are all remapped from the same remapping layer of $\gN_s$. This remapping strategy prevents the huge cost of ImageNet pre-training involved in the search space, i.e. the super network in differentiable NAS.
    
\begin{table*}[h!]
    \centering
    \caption{Semantic segmentation results on the Cityscapes validation set. ``OS": output stride, the spatial resolution ratio of the input image to the backbone output. ``iters": the number of total training iterations. The result of DPC in the brackets is our re-implemented version under the same settings as \name. The MAdds of the models are computed with the $1024 \times 2048$ input resolution. \rev{We report the \name\ results as $mean_{\pm std}$ for 3 independent runs (the same as other tables).}}
    \label{tab: seg}
    \begin{threeparttable}
        \begin{tabular}{=l | +c | +c | +c | +c | +c | +c}
        \hline
        \multicolumn{2}{l|}{\textbf{Method}} & \textbf{OS} &\textbf{iters} & \textbf{Params} & \textbf{MAdds} & \textbf{mIOU(\%)} \\
        \hline
        MobileNetV2~\cite{sandler2018mobilenetv2} & \multirow{4}*{DeepLabv3} &\multirow{4}*{16}& \multirow{4}*{100K} & 2.57M & 24.52B & 75.5 \\
        DPC~\cite{DBLP:conf/nips/ChenCZPZSAS18} & && & 2.51M & 24.69B & 75.4(75.7)\\
        FNA~\cite{Fang*2020Fast} & & && 2.47M & 24.17B & 76.6\\
        \rowstyle{\color{revcolor}}
        \name & & && 2.37$_{\pm 0.08}$M & 24.3$_{\pm 0.30}$B & \textbf{77.0}$_{\pm 0.14}$\\
        \hline
        Auto-DeepLab-S~\cite{liu2019auto} & \multirow{5}*{DeepLabv3+} & 8 & 500K & 10.15M & 333.25B& 75.2\\
        FNA~\cite{Fang*2020Fast} & & 16& 100K & 5.71M & 210.11B & 77.2 \\
        \name & & 16 & 100K & 5.71M & 210.11B & 78.2 \\
        FNA~\cite{Fang*2020Fast} & & 8 & 100K & 5.71M & 313.87B & 78.0 \\
        \name & & 8 & 100K & 5.71M & 313.87B & \textbf{78.4} \\
        \hline
        \end{tabular}
    \end{threeparttable}
\end{table*}

\begin{table*}[h!]
    \centering
    \caption{Comparison of computational cost on the semantic segmentation task. ``ArchAdapt": architecture adaptation. ``ParamAdapt": parameter adaptation. ``GHs": GPU hours. $\ast$ denotes the computation cost computed under our reproducing settings. $\dagger$ denotes the cost estimated according to the description in the original paper~\cite{liu2019auto}.}
    \label{tab: seg cost}
    \begin{threeparttable}
        \begin{tabular}{l c c c}
        \toprule
        \textbf{Method} & \textbf{Total Cost} & \textbf{ArchAdapt Cost} & \textbf{ParamAdapt Cost} \\
        \midrule
        DPC~\cite{DBLP:conf/nips/ChenCZPZSAS18} & 62.2K GHs & 62.2K GHs & 30.0$^\ast$ GHs\\
        Auto-DeepLab-S~\cite{liu2019auto} & 244.0 GHs & 72.0 GHs & 172.0$^\dagger$ GHs\\
        \name & 35.8 GHs & 1.4 GHs & 34.4 GHs\\
        \bottomrule
    \end{tabular}
    \end{threeparttable}
\end{table*}

    \subsubsection{Architecture Adaptation}
    We start the differentiable NAS process with the expanded super network directly on the target task, \ie, semantic segmentation, object detection and human pose estimation in our experiments. As NAS works commonly do, we split a part of data from the original training dataset as the validation set. In the preliminary search epochs, as the operation weights are not sufficiently trained, the architecture parameters cannot be updated towards a clear and correct direction. We first train operation weights of the super network for some epochs on the training dataset, which is also mentioned in some previous differentiable NAS works~\cite{fbnet,liu2019auto}. After the weights get sufficiently trained, we start alternating the optimization of operation weights $w$ and architecture parameters $\alpha$. Specifically, we update $w$ on the training dataset by computing ${\partial \mathcal{L}} / {\partial w}$ and optimize $\alpha$ on the validation dataset with ${\partial \mathcal{L}} / {\partial \alpha}$. To control the computation cost (MAdds in our experiments) of the searched network, we define the loss function as follows.
    \begin{equation}
        \mathcal{L} = \mathcal{L}_{task} + \lambda \lg cost\text{,}
    \label{eq: loss}
    \end{equation}
    where $\lambda$ in the second term controls the magnitude of the MAdds optimization. The $cost$ term during search is computed as
    \begin{equation}
        \begin{aligned}
            cost^{(l)} &= \sum_{o \in O} \frac{exp(\alpha^{(l)}_{o})}{\sum_{o' \in O} exp(\alpha^{(l)}_{o'})}cost_o^{(l)}\text{,}\\
            cost &= \sum_l cost^{(l)}\text{,}
        \end{aligned}
    \end{equation}
    where $cost_o^{(l)}$ is obtained by measuring the cost of operation $o$ in layer $l$, $cost^{(l)}$ is the total cost of layer $l$ which is computed by a weighted-sum of all operation costs and $cost$ is the total cost of the network obtained by summing the cost $cost^{(l)}$ of all the layers.
    To accelerate the search process and decouple the parameters of different sub-networks, we only sample one path in each iteration according to the distribution of architecture parameters for operation weight updating. As the search process terminates, we use the architecture parameters $\alpha$ to derive the target architecture $\mathit{Arch}_t$. The final operation type in each searched layer is determined as the one with the maximum architecture parameter $\alpha$.

    \subsubsection{Parameter Adaptation}
    We obtain the target architecture $\mathit{Arch}_t$ from architecture adaptation. To accommodate the new tasks, the target architecture becomes different from that of the seed network $\gN_s$ (which is primitively designed for the image classification task). Unlike conventional training strategy, we discard the cumbersome pre-training process of $\mathit{Arch}_t$ on ImageNet. We remap parameters of $\gN_{sup}$ to $\mathit{Arch}_t$ before parameter adaptation. As shown in Fig.~\ref{fig: framework}, the parameters of every searched layer in $\mathit{Arch}_t$ are remapped from the operation with the same type in the corresponding layer in $\gN_{sup}$. As the shape of the parameters is the same for the same operation type, the remapping process here can be performed as a pure collection manner. All the other layers in $\mathit{Arch}_t$, including the input convolution and the head part of the network \etc, are directly remapped from $\gN_{sup}$ as well. In our former conference version~\cite{Fang*2020Fast}, the parameters of $\mathit{Arch}_t$ are remapped from the seed network $\gN_s$. We find that performing parameter remapping from $\gN_{sup}$ can achieve better performance than from $\gN_s$. We further study the remapping mechanism for parameter adaptation in experiments (Sec.~\ref{subsec: map_mech}). With parameter remapping on $\mathit{Arch}_t$ finished, we fine-tune the parameters of $\mathit{Arch}_t$ on the target task and obtain the final target network $\gN_t$.

% --------------------------------Section Partition------------------------------- %
\section{Experiments}
\label{sec: Experiments}
In this section, we first select the ImageNet pre-trained model MobileNetV2~\cite{sandler2018mobilenetv2} as the seed network and apply our \name\ method on three common computer vision tasks, \ie, semantic segmentation in Sec.~\ref{subsec: seg}, object detection in Sec.~\ref{subsec: det} and human pose estimation in Sec.~\ref{subsec: pose}. 
% Then we analyze the changes of the adapted architectures to the seed network in Sec.~\ref{subsec: anal_arch}. 
We further implement \name\ on more network types to demonstrate the generalization ability, including ResNets~\cite{he2016deep} in Sec.~\ref{subsec: resnet} and NAS networks in Sec.~\ref{subsec: nasnets}. 

\subsection{Network Adaptation on Semantic Segmentation}
\label{subsec: seg}
The semantic segmentation experiments are conducted on the Cityscapes~\cite{DBLP:conf/cvpr/CordtsORREBFRS16} dataset. In the architecture adaptation process, we map the seed network to the super network, which is used as the backbone of DeepLabv3~\cite{deeplab-v3}. The whole search process is conducted on a single V100 GPU and takes only 1.4 hours in total. In the parameter adaptation process, we remap the parameters of the super network to the target architecture obtained in the aforementioned architecture adaptation. The whole parameter adaptation process is conducted on $4$ TITAN-Xp GPUs and takes $100$K iterations, which costs only $8.5$ hours in total. All the other searching and training hyper-parameters are the same as that in~\cite{Fang*2020Fast}.

\begin{table*}[h]
    \centering
    \caption{Object detection results on MS-COCO. The MAdds are computed with $1088 \times 800$ input images for RetinaNet and $320 \times 320$ for SSDLite.}
    \label{tab: det}
    \begin{threeparttable}
        \begin{tabular}{=l | +c | +c | +c | +c }
        \hline
        \multicolumn{2}{l|}{\textbf{Method}} & \textbf{Params} & \textbf{MAdds} & \textbf{mAP(\%)} \\
        \hline
        ShuffleNetV2-20~\cite{DBLP:journals/corr/abs-1903-10979} & \multirow{5}*{RetinaNet} & 13.19M &    132.76B & 32.1 \\
        MobileNetV2~\cite{sandler2018mobilenetv2} & & 11.49M & 133.05B & 32.8\\
        DetNAS~\cite{DBLP:journals/corr/abs-1903-10979} & & 13.41M & 133.26B & 33.3\\
        FNA~\cite{Fang*2020Fast} & & 11.73M & 133.03B & 33.9 \\
        \rowstyle{\color{revcolor}}
        \name & & 11.90$_{\pm 0.01}$M & 132.86$_{\pm 0.15}$B & \textbf{34.7}$_{\pm 0.05}$ \\
        \hline
        MobileNetV2~\cite{sandler2018mobilenetv2} &\multirow{4}*{SSDLite}&4.3M &0.8B &22.1 \\
        Mnasnet-92~\cite{MnasNet} & & 5.3M &1.0B &22.9 \\
        FNA~\cite{Fang*2020Fast} & & 4.6M &0.9B & 23.0 \\
        \rowstyle{\color{revcolor}}
        \name & & 4.3$_{\pm 0.09}$M & 0.9$_{\pm 0.01}$B & \textbf{23.9}$_{\pm 0.09}$ \\
        \hline
    \end{tabular}
    \end{threeparttable}
\end{table*}

\begin{table*}[h]
    \centering
    \caption{Comparison of computational cost on the object detection task. All our experiments on object detection are conducted on TITAN-Xp GPUs. ``GDs": GPU days.}
    \label{tab: det cost}
    \begin{threeparttable}
        \begin{tabular}{l| c| c |c| c| c| c}
        \hline
        \multirow{2}*{\textbf{Method}} & \multirow{2}*{\textbf{Total Cost}} & \multicolumn{3}{c|}{\textbf{Super Network}} & \multicolumn{2}{c}{\textbf{Target Network}} \\
        \cline{3-7}
        & & Pre-training &Finetuning &Search &Pre-training & Finetuning \\
        \hline
        DetNAS~\cite{DBLP:journals/corr/abs-1903-10979} &68 GDs &12 GDs &12 GDs &20 GDs &12 GDs &12 GDs \\
        \name\ (RetinaNet) & 8.5 GDs & - & - & 5.3 GDs & - &3.2 GDs \\
        \name\ (SSDLite) & 21.0 GDs & - & - & 5.7 GDs & - & 15.3 GDs \\
        \hline
        \end{tabular}
    \end{threeparttable}
\end{table*}

% \subsubsection{Experimental Results}
\rev{To obtain more convincing results, we independently run the whole adaptation process three times with different seeds and report the mean and std.} The semantic segmentation results are shown in Tab.~\ref{tab: seg}. The \name\ network achieves a $77.0\%$ mean mIOU on Cityscapes with DeepLabv3~\cite{deeplab-v3}, $1.5\%$ mIOU better than the handcrafted seed MobileNetV2~\cite{sandler2018mobilenetv2} with fewer MAdds. Compared with the NAS method DPC~\cite{DBLP:conf/nips/ChenCZPZSAS18} (with MobileNetV2 as the backbone) which searches a multi-scale module for semantic segmentation, \name\ gets $1.3\%$ mIOU promotion with $0.39$B fewer MAdds. For fair comparison with Auto-DeepLab~\cite{liu2019auto} which searches the backbone architecture on DeepLabv3 and retrains the searched network on DeepLabv3+~\cite{chen2018encoder}, we adapt the parameters of the target architecture $\mathit{Arch}_t$ to the DeepLabv3+ framework. Comparing with Auto-DeepLab-S, \name\ achieves far better mIOU with fewer MAdds, Params and training iterations. With the output stride of 16, \name\ promotes the mIOU by $3.0\%$ with only $63.0\%$ MAdds of Auto-DeepLab-S. With the improved remapping mechanism for parameter adaptation, \name\ achieves better performance than our former version~\cite{Fang*2020Fast}. We compare the computation cost in Tab.~\ref{tab: seg cost}. With the remapping mechanism, \name\ greatly decreases the computation cost for adaptation, only taking 35.8 GPU hours, $1737\times$ less than DPC and $6.8\times$ less than Auto-DeepLab.

\subsection{Network Adaptation on Object Detection}
\label{subsec: det}
We further implement our \name\ method on object detection tasks. We adapt the MobileNetV2 seed network to two commonly used detection systems, RetinaNet~\cite{lin2017focal} and SSDLite~\cite{liu2016ssd, sandler2018mobilenetv2}, on the MS-COCO dataset~\cite{DBLP:conf/eccv/LinMBHPRDZ14}. For RetinaNet, the short side of the input image is resized to 800 while the maximum long side is set as 1088 to obtain a larger batch size for search. The whole architecture search process takes $14$ epochs, $16$ hours on 8 TITAN-Xp GPUs with the batch size of 16 and the whole parameter fine-tuning takes 12 epochs, about $10$ hours on 8 TITAN-Xp GPUs with 32 batch size. For SSDLite, the search process takes $30$ epochs in total, $17$ hours on $8$ TITAN-Xp GPUs with $128$ batch size, and the parameter adaptation takes $60$ epochs, $46$ hours on $8$ TITAN-Xp GPUs with $512$ batch size. The other searching and training hyper-parameters are following~\cite{Fang*2020Fast}.
We show the results on the MS-COCO dataset in Tab.~\ref{tab: det}. For the RetinaNet framework, compared with two manually designed networks, ShuffleNetV2-10~\cite{DBLP:journals/corr/abs-1807-11164, DBLP:journals/corr/abs-1903-10979} and MobileNetV2~\cite{sandler2018mobilenetv2}, \name\ achieves higher mAP with similar MAdds. Compared with DetNAS~\cite{DBLP:journals/corr/abs-1903-10979} which searches the backbone of the detection network, \name\ achieves $1.4\%$ higher mAP with $1.5$M fewer Params and $0.4$B fewer MAdds. As shown in Tab.~\ref{tab: det cost}, our total computation cost is only $12.5\%$ of DetNAS on RetinaNet. For SSDLite in Tab.~\ref{tab: det}, \name\ surpasses both the manually designed network MobileNetV2 and the NAS-searched network MnasNet-92~\cite{MnasNet}, while MnasNet takes around 3.8K GPU days to search for the backbone network on ImageNet~\cite{imagenet}. The total computation cost of MnasNet is far larger than ours and is unaffordable for most researchers or engineers. The specific cost \name\ takes on SSDLite is shown in Tab.~\ref{tab: det cost}. It is difficult to train the small network due to the simplification~\cite{liu2019training}. Therefore, experiments on SSDLite need longer training schedules and take larger computation cost than RetinaNet. The experimental results further demonstrate the efficiency and effectiveness of direct adaptation on the target task with parameter remapping and architecture search.

\begin{table}[t]
    \centering
    \caption{Human pose estimation results on the MPII validation set.}
    \begin{threeparttable}
        \begin{tabular}{=l+l+l+l}
        \toprule
        \textbf{Method} & \textbf{Params} & \textbf{MAdds} & \textbf{PCKh@0.5} \\
        \midrule
        MobileNetV2 & 5.23M & 6.09B & 85.9 \\
        \rowstyle{\color{revcolor}}
        \name & 5.25$_{\pm 0.16}$M & 6.16$_{\pm 0.02}$B & \textbf{87.0}$_{\pm 0.10}$ \\
        \bottomrule
        \end{tabular}
    \label{tab: pose}
    \end{threeparttable}
\end{table}

\subsection{Network Adaptation on Human Pose Estimation}
\label{subsec: pose}
We apply \name\ on the human pose estimation task. The experiments are performed on the MPII dataset~\cite{andriluka20142d} with the SimpleBaseline framework~\cite{xiao2018simple}. MPII dataset contains around 25K images with about 40K people. For the search process in architecture adaptation, we randomly sample $20\%$ data from the original training set as the validation set for architecture parameter optimization. The other $80\%$ data is used as the training set for search. For architecture parameters, we use the Adam optimizer~\cite{DBLP:journals/corr/KingmaB14} with a fixed learning rate of $3 \times 10^{-3}$ and $0$ weight decay. We set $\lambda$ in Eq.~\ref{eq: loss} as $3 \times 10^{-5}$ for MAdds optimization. The input image is cropped and then resized to $256 \times 256$ following the standard training settings~\cite{xiao2018simple,sun2019deep}. The batch size is set as 32. All the other training hyper-parameters are the same as SimpleBaseline. The search process takes $140$ epochs in total and the architecture parameter updating starts after 70 epochs. For parameter adaptation, we use the same training settings as SimpleBaseline. PCKh@0.5~\cite{andriluka20142d} is used as the evaluation metric.

The architecture adaptation takes 16 hours in total on only one TITAN X GPU and parameter adaptation takes 5.5 hours on one TITAN X GPU. The total computation cost is 21.5 GPU hours. As shown in Tab.~\ref{tab: pose}, \name\ promotes the PCKh@0.5 by $1.1$ with similar model MAdds. As we aim at validating the effectiveness of \name\ on networks, we do not tune the training hyper-parameters and just follow the default ResNet-50~\cite{he2016deep} training settings in SimpleBaseline for both MobileNetV2 and the \name\ network training.

\begin{figure}[!t]
    \centering
    \includegraphics[width=0.78\columnwidth]{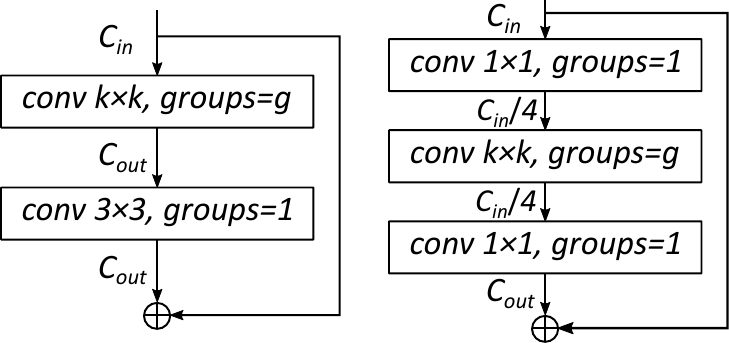}
    \caption{The searchable ResNet blocks in \name, including the basic block (\emph{left}) and the bottleneck block (\emph{right}). The first convolution in the basic block and the middle convolution in the bottleneck block are searchable, while the kernel size is denoted as ``$k \times k$" and the group number is denoted as ``$g$".}
    \label{fig: res_block}
\end{figure}

\begin{table}[!t]
    \centering
    \caption{Optional block types in ResNet search space. The block type ``k3g1" is equivalent to the original basic block or bottleneck block in ResNet~\cite{he2016deep}. The type with the group number of 1 represents the plain convolution.}
    \begin{tabular}{c|c|c}
    \hline
    \textbf{block type} & \textbf{kernel size} & \textbf{group number}\\
    \hline
    k3g1 & 3 & 1\\
    k5g2 & 5 & 2\\
    k5g4 & 5 & 4\\
    k7g4 & 7 & 4\\
    k7g8 & 7 & 8\\
    \hline
    \end{tabular}
    \label{tab: res_space}
\end{table}

\begin{table}[!t]
    \centering
    \caption{Object detection results of RetinaNet on MS-COCO with ResNets~\cite{he2016deep} as the seed networks.}
    \begin{threeparttable}
        \begin{tabular}{l cccc }
        \toprule
        \textbf{Model} & \textbf{Params} & \textbf{MAdds} & \textbf{mAP(\%)} \\
        \midrule
        ResNet-18 & 21.41M & 160.28B & 32.1 \\
        \name & 20.66M & 159.64B & \textbf{32.9} \\
        \midrule
        ResNet-50 & 37.97M & 202.84B & 35.5 \\
        \name & 36.27M & 200.33B & \textbf{36.8} \\
        \bottomrule
    \end{tabular}
    \end{threeparttable}
    \label{tab: res_seed}
\end{table}

\subsection{Network Adaptation on ResNet}
\label{subsec: resnet}
To evaluate the generalization ability on different network types, we perform our method on ResNets~\cite{he2016deep}, including ResNet-18 and ResNet-50. As ResNets are composed of plain convolutions, kernel size enlargement will cause huge MAdds increase. We propose to search for diverse kernel sizes in ResNets without much MAdds increase by introducing grouped convolutions~\cite{krizhevsky2012imagenet}. The searchable ResNet blocks are shown in Fig.~\ref{fig: res_block}. We allow the first convolution in the basic block and the second convolution in the bottleneck block to be searched. All the optional block types in the designed ResNet search space are shown in Tab.~\ref{tab: res_space}. As the kernel size enlarges, we set more groups in the convolution block to maintain the MAdds. 

We perform the adaptation on ResNet-18 and -50 to the RetinaNet~\cite{lin2017focal} framework. For ResNet-18, the input image for search is resized to ones with the short side to $800$ and the long side not exceeding $1333$ (shortly denoted as $(1333, 800)$ in MMDetection~\cite{chen2019mmdetection}). The SGD optimizer for operation weights is used with $5\times10^{-5}$ weight decay and $0.02$ initial learning rate. $\lambda$ in Eq.~\ref{eq: loss} is set as $0.08$. All the other search and training settings are the same as the MobileNetV2 experiments on RetinaNet. The total adaptation cost is only $6$ GPU days, including $10.5$ hours on $8$ TITAN-Xp GPUs for search and $7.5$ hours on $8$ GPUs for parameter adaptation. For ResNet-50, the batch size is set as $8$ in total for search. The input image is also resized to $(1333, 800)$. For the SGD optimizer, the initial learning rate is $0.01$ and the weight decay is $1\times10^{-4}$. The other hyper-parameters for search are the same as that for ResNet-18. For the training in parameter adaptation, we first recalculate the running statistics of BN for $200$ iterations with the synchronized batch normalization across GPUs (SyncBN). Then we freeze the BN layers\footnote[1]{Freezing BN means using the running statistics of BN during training and not updating the BN parameters. It is implemented as \emph{.eval()} in PyTorch~\cite{paszke2017automatic}.} and train the target architecture on MS-COCO using the same hyper-parameters as ResNet-50 training in MMDetection. The architecture adaptation takes $29.5$ hours and parameter adaptation takes $12.5$ hours on $8$ TITAN-Xp GPUs, $14$ GPU days in total. The results are shown in Tab.~\ref{tab: res_seed}. Compared with the original ResNet-18 and -50, \name\ can further promote the mAP by $0.8$ and $1.3$ with fewer Params and MAdds.

\begin{table}[th]
    \centering
    \caption{Search space with NAS networks as the seed networks. ``FB" denotes the network FBNet-C and ``Proxy" denotes Proxyless (mobile). The other abbreviations are the same as Tab.~\ref{tab: space_mbv2}.}
    \begin{tabular}{c|c|c|c|c}
    \hline
    \multirow{2}*{\textbf{Block}} & \multicolumn{2}{c|}{\textbf{chs}} & \multirow{2}*{\textbf{n}} & \multirow{2}*{\textbf{s}} \\
    \cline{2-3}
    & FB & Proxy &&\\
    \hline
    $3 \times 3$ Conv & 16  & 32  & 1 & 2 \\
    MBConv(k3e1)      & 16  & 16  & 1 & 1 \\
    SBlock            & 24  & 32  & 4 & 2 \\
    SBlock            & 32  & 40  & 4 & 2 \\
    SBlock            & 64  & 80  & 4 & 2 \\
    SBlock            & 112 & 96  & 4 & 1 \\
    SBlock            & 184 & 192 & 4 & 2 \\
    SBlock            & 352 & 320 & 1 & 1 \\
    \hline    
    \end{tabular}
    \label{tab: space_nas}
\end{table}

\begin{table}[th]
    \centering
    \caption{Object detection results of RetinaNet on MS-COCO with NAS networks as the seed networks.}
    \begin{threeparttable}
        \begin{tabular}{l cccc }
        \toprule
        \textbf{Model} & \textbf{Params} & \textbf{MAdds} & \textbf{mAP(\%)} \\
        \midrule
        MobileNetV2~\cite{sandler2018mobilenetv2} & 11.49M & 133.05B & 32.8\\
        \midrule
        FBNet-C~\cite{fbnet} & 12.65M & 134.17B & 34.9 \\
        \name & 12.51M & 134.20B & \textbf{35.5} \\
        \midrule
        Proxyless (mobile)~\cite{cai2018proxylessnas} & 12.07M & 133.42B & 34.6 \\
        \name & 12.10M & 133.23B & \textbf{35.3} \\
        \bottomrule
    \end{tabular}
    \end{threeparttable}
    \label{tab: nas_seed}
\end{table}

\subsection{Network Adaptation on NAS networks}
\label{subsec: nasnets}
Our proposed parameter remapping paradigm can be implemented on various types of networks. We further apply \name\ on two popular NAS networks, \ie, FBNet-C~\cite{fbnet} and Proxyless (mobile)~\cite{cai2018proxylessnas}. The search space is constructed as Tab.~\ref{tab: space_nas} shows. FBNet and ProxylessNAS search for architectures on the ImageNet classification task. To compare with the seed networks FBNet-C and Proxyless (mobile), we re-implement the two NAS networks and deploy them on the RetinaNet~\cite{lin2017focal} framework. Then we train them on the MS-COCO~\cite{COCO} dataset with the ImageNet pre-trained parameters using the same training hyper-parameters as ours. The results are shown in Tab.~\ref{tab: nas_seed}. Though the NAS networks already achieve far better performance than handcrafted MobileNetV2 on the detection task, our \name\ networks further promote the mAP which cost similar MAdds with the NAS seed networks. This experiment demonstrates that \name\ can not only promote the performance of manually designed networks, but also improve the NAS networks which are not searched on the target task. In real applications, if there is a demand for a new task, \name\ helps to adapt the network with a low cost, avoiding cumbersome cost for extra pre-training and huge cost for searching from scratch. We visualize the architectures in Fig.~\ref{fig: nas_networks} of Appendix. Similar to the MobileNetV2 seed network, adaptation on the ImageNet NAS networks decreases the layer numbers in the heavy stage 2 and introduces more large-kernel convolutions.

\rev{\section{Ablation Study}
\label{sec: ablat}
In this section, we perform a series of ablation studies to demonstrate the effectiveness of our method. We first study the remapping mechanism for parameter adaptation in Sec.~\ref{subsec: map_mech} by comparing and analyzing two remapping mechanisms. In Sec.~\ref{subsec: effect PR}, we evaluate the effectiveness of parameter remapping for the two adaptation stages. We compare the proposed remapping method with Net2Net~\cite{chen2015net2net} in Sec.~\ref{subsec: comp_net2net}, and study several different remapping strategies in Sec.~\ref{subsec: studyPR}. Then in Sec.~\ref{subsec: randsearch}, random search experiments are performed to demonstrate our method can be used as a NAS-method agnostic mechanism, and random sampling is compared to evaluate the improvements brought by architecture search.
}

\begin{figure*}[thbp]
    \centering
    \subfigure[RetinaNet]{
    \begin{minipage}[b]{0.32\linewidth}
    \includegraphics[width=1\linewidth]{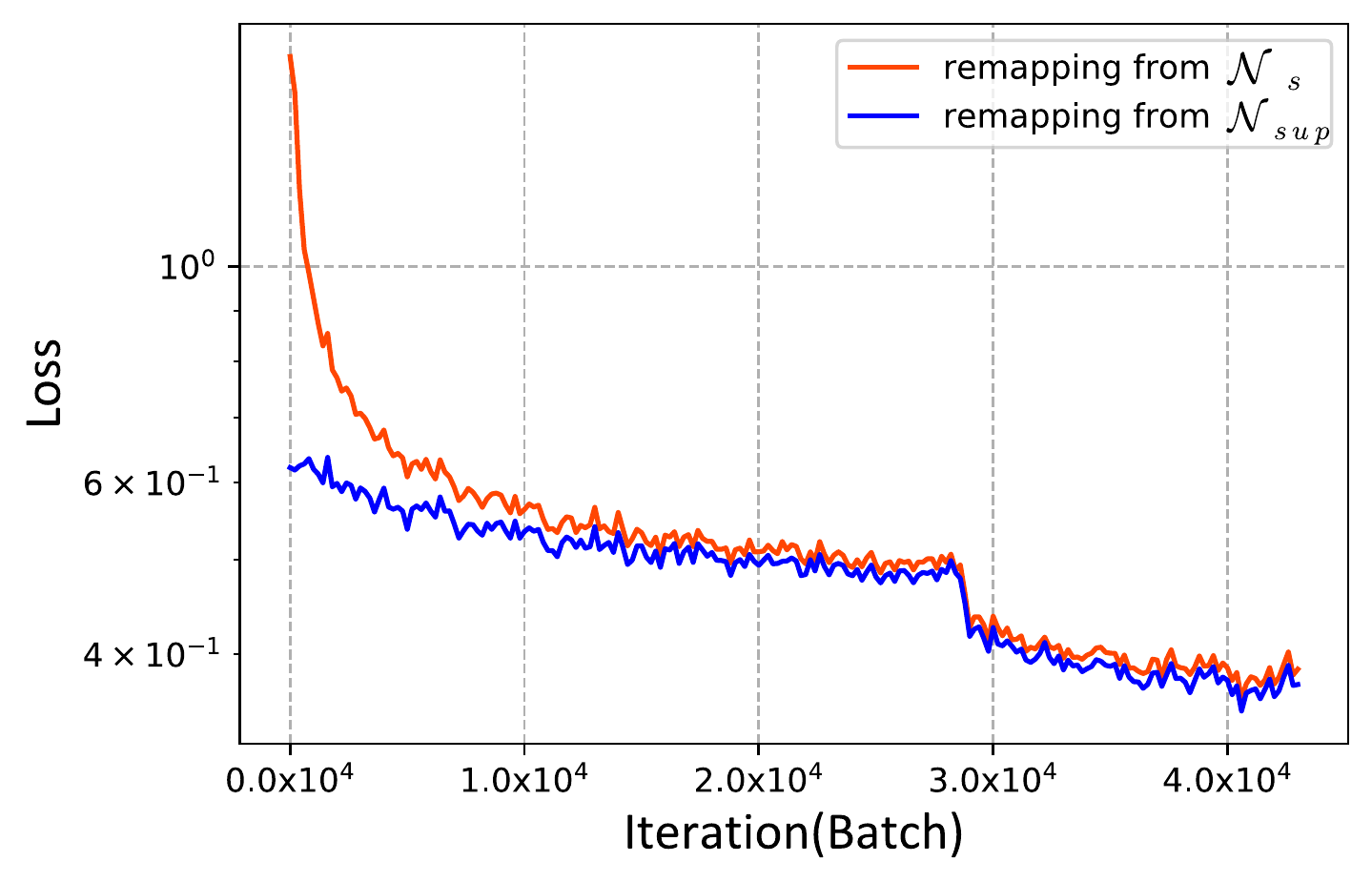}\vspace{4pt}
    \includegraphics[width=1\linewidth]{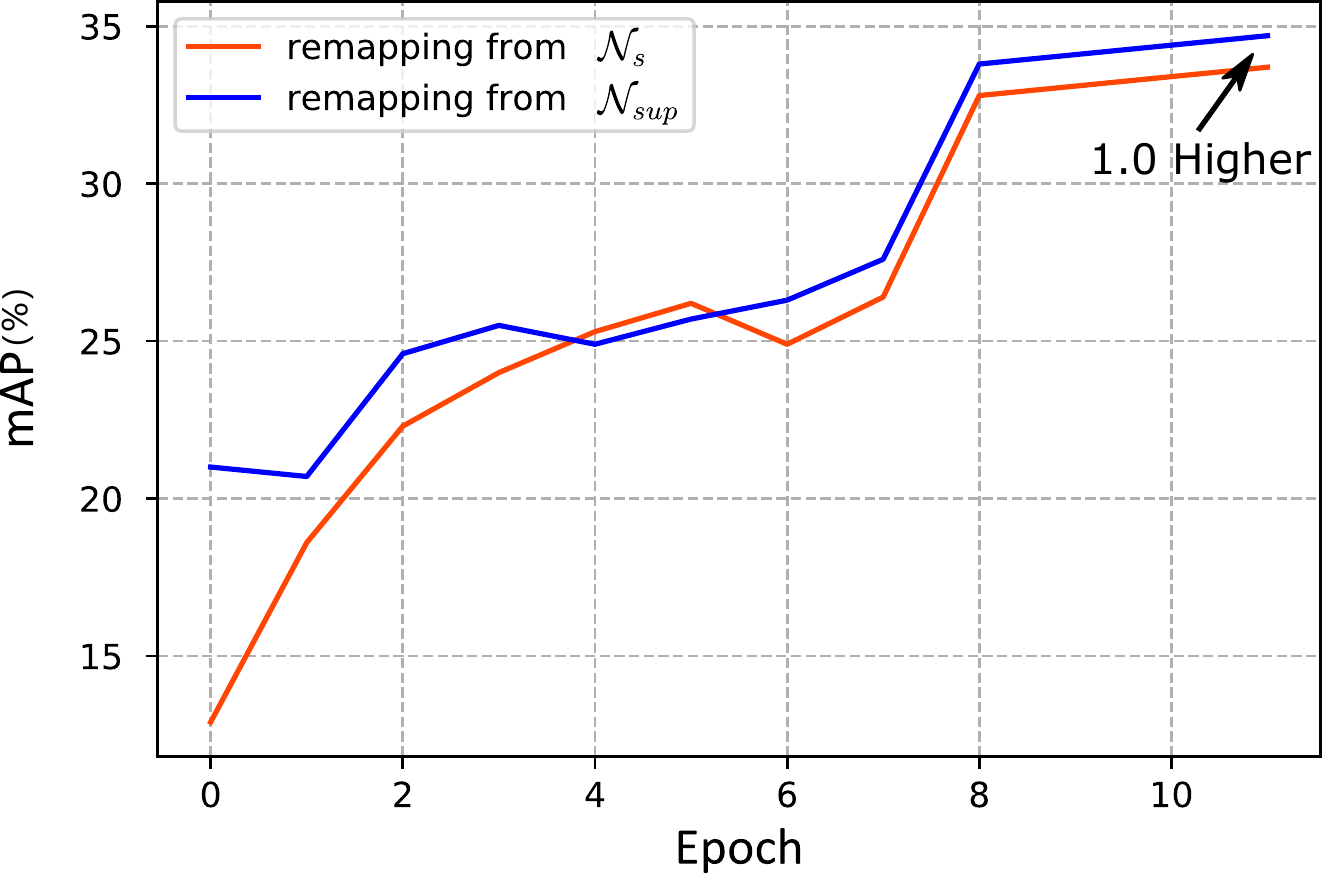}
    \end{minipage}}
    \subfigure[RetinaNet ($2\times$)]{
    \begin{minipage}[b]{0.32\linewidth}
    \includegraphics[width=1\linewidth]{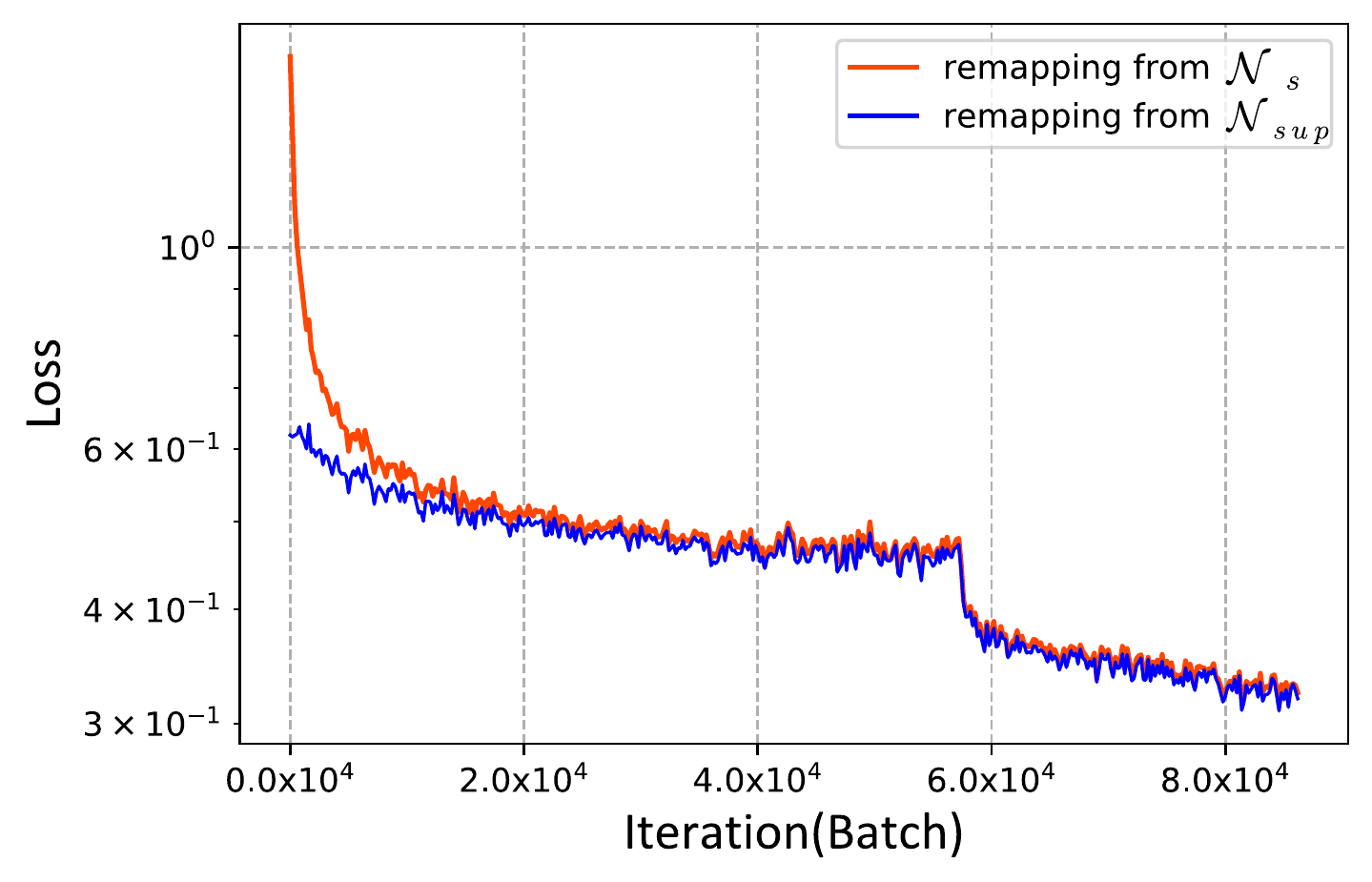}\vspace{4pt}
    \includegraphics[width=1\linewidth]{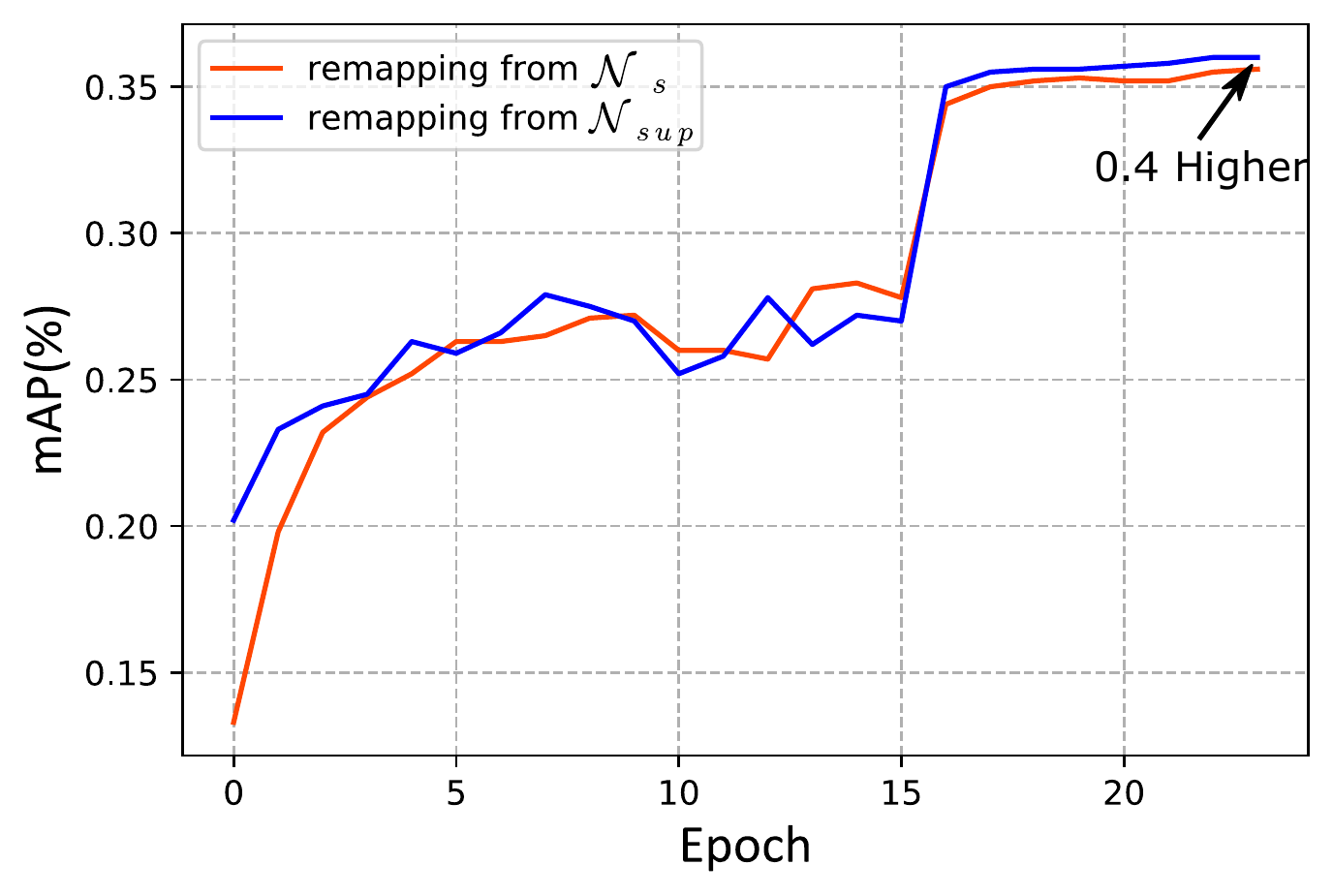}
    \end{minipage}}
    \subfigure[SSDLite]{
    \begin{minipage}[b]{0.32\linewidth}
    \includegraphics[width=1\linewidth]{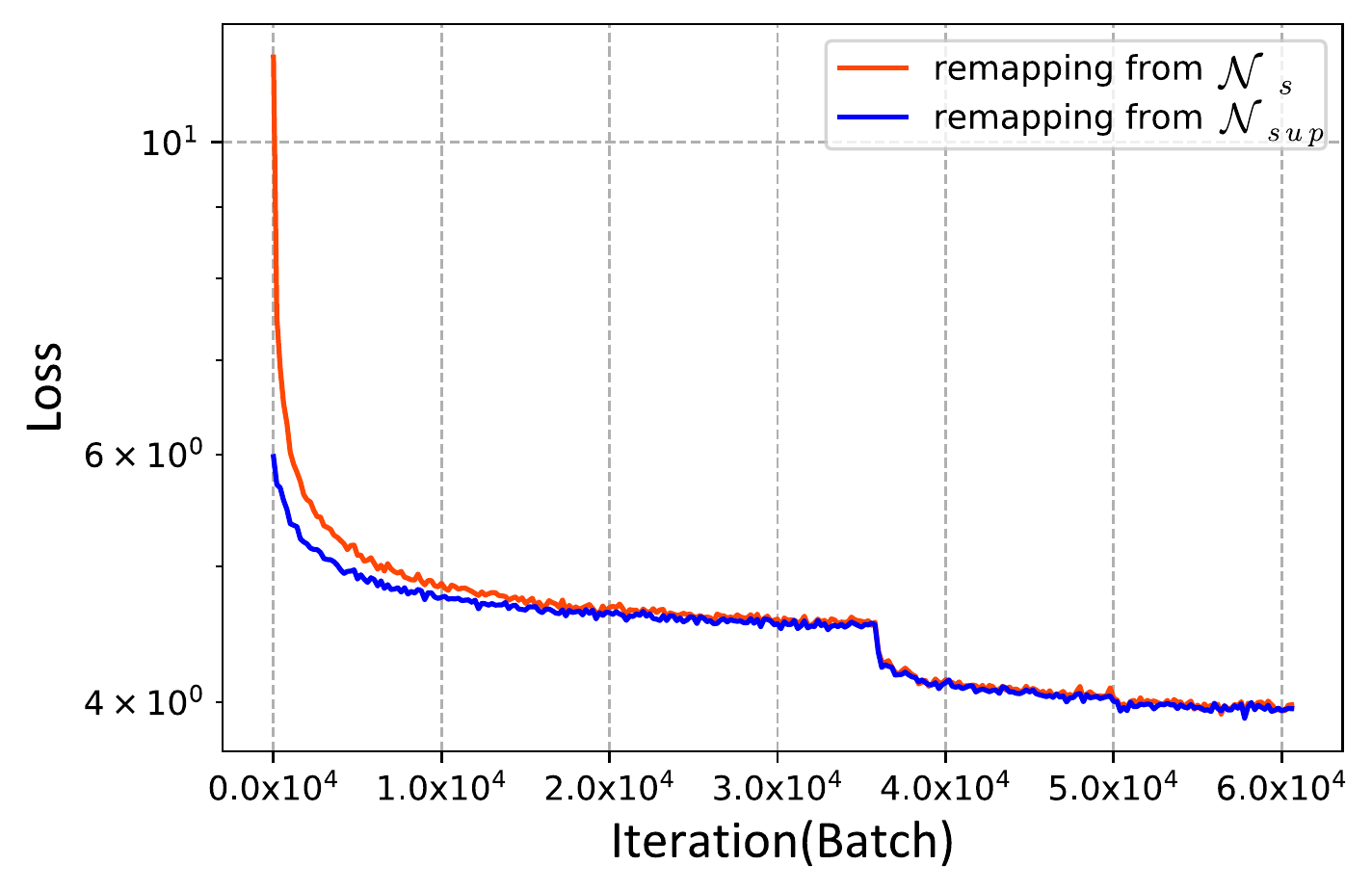}\vspace{4pt}
    \includegraphics[width=1\linewidth]{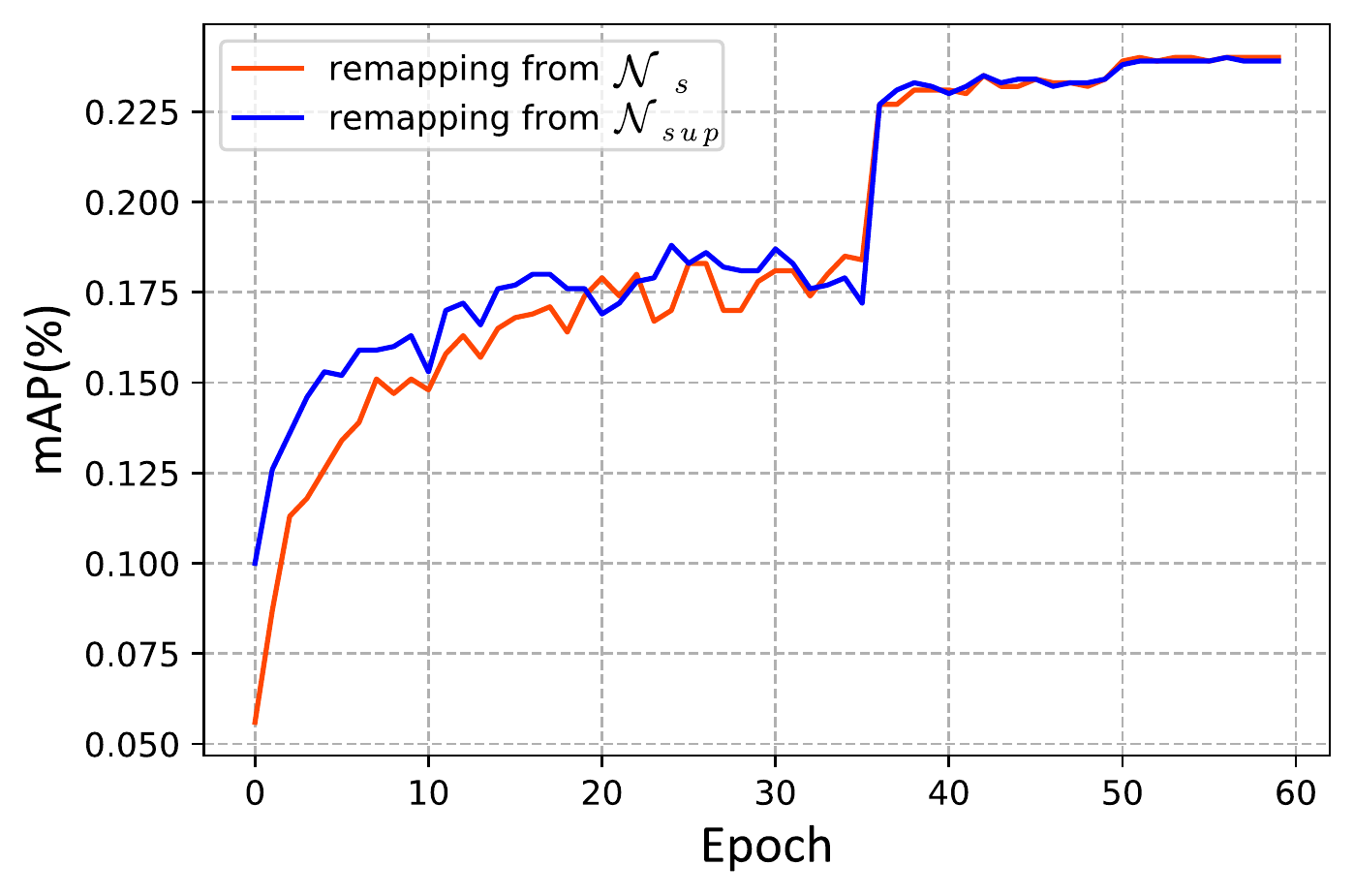}
    \end{minipage}}
    \centering
    \caption{Training loss (upper) and mAP (bottom) comparisons between two remapping mechanisms on MS-COCO, \ie remapping parameters from the seed network $\gN_s$ (\emph{red}) and from the super network $\gN_{sup}$ (\emph{blue}). Remapping from the super network greatly accelerates the training convergence in early epochs and achieves higher performance. With the training schedule lengthened ($2\times$), the performance gap between two remapping mechanisms narrows. As SSDLite training takes long epochs, two remapping mechanisms achieve the same results.}
    \label{fig: loss/map}
\end{figure*}

\subsection{Study the Remapping Mechanism for Parameter Adaptation}
\label{subsec: map_mech}
In our preliminary version~\cite{Fang*2020Fast}, with the target architecture $\mathit{Arch}_t$ obtained by architecture adaptation, we remap the parameters of the seed network $\gN_s$ to the target architecture $\mathit{Arch}_t$ for latter parameter adaptation. As we explore the mechanism of parameter remapping, we find that parameters remapped from the super network $\gN_{sup}$ can bring further performance promotion for parameter adaptation. However, the batch normalization (BN) parameters during search may cause instability and damage the training performance of the sub-architectures in the super network. The parameters of BN are usually disabled during search in many differentiable/one-shot NAS methods~\cite{liu2018darts,Understanding}. We open the BN parameter updating in the search process, including learnable affine parameters and global mean/variance statistics, so as to completely use parameters from $\gN_{sup}$ for parameter adaptation. Experiments show that BN parameters updating causes little effect on the search performance. 

\begin{table}[t]
    \centering
    \caption{Comparison results using different remapping mechanisms with RetinaNet on MS-COCO. ``from \emph{seed}" denotes parameters of $\mathit{Arch}_t$ are remapped from the seed network $\gN_s$. ``from \emph{sup}" denotes parameters of $\mathit{Arch}_t$ are remapped from the super network $\gN_{sup}$. ``$2\times$" denotes the longer training schedule for RetinaNet, \ie 24 epochs in MMDetection~\cite{chen2019mmdetection}.}
    \label{tab: remap_mech_det}
    \begin{threeparttable}
        \begin{tabular}{l | c | c | c | l }
        \hline
        \multicolumn{2}{l|}{\textbf{Method}} & \textbf{Params} & \textbf{MAdds} & \textbf{mAP(\%)} \\
        \hline
        from \emph{seed} & \multirow{4}*{RetinaNet} & 11.91M & 132.99B & 33.7 \\
        from \emph{sup} &  & 11.91M & 132.99B & 34.7$_{\uparrow1.0}$ \\
        \cline{0-0} \cline{3-5}
        from \emph{seed} ($2\times$) &  & 11.91M & 132.99B & 35.6 \\
        from \emph{sup} ($2\times$) &  & 11.91M & 132.99B & 36.0$_{\uparrow0.4}$ \\
        \hline
        from \emph{seed} & \multirow{2}*{SSDLite} & 4.4M & 0.9B & 24.0 \\
        from \emph{sup} &  & 4.4M & 0.9B & 24.0 \\
        \hline
    \end{tabular}
    \end{threeparttable}
\end{table}

\begin{table}[t]
    \centering
    \caption{Comparison results with different remapping mechanisms for semantic segmentation on Cityscapes. The experiments are performed on DeepLabv3. All the abbreviation definitions are the same as Tab.~\ref{tab: remap_mech_det}.}
    \label{tab: remap_mech_seg}
    \begin{threeparttable}
        \begin{tabular}{l | c | c | l }
        \hline
        \textbf{Method} & \textbf{Params} & \textbf{MAdds} & \textbf{mIOU(\%)} \\
        \hline
        from \emph{seed} & 2.47M & 24.17B & 76.6 \\
        from \emph{sup}  & 2.47M & 24.17B & 77.1$_{\uparrow0.5}$ \\
        \hline
    \end{tabular}
    \end{threeparttable}
\end{table}

\begin{table*}[thbp]
    \centering
    \caption{Effectiveness evaluation of parameter remapping. The experiments are conducted with DeepLabv3 on Cityscapes.  ``Remap": parameter remapping from the seed network. ``ArchAdapt": architecture adaptation. ``RemapSuper": parameter remapping from the super network. ``ParamAdapt": parameter adaptation. ``RandInit": random initialization. ``Pretrain": ImageNet pre-training of the target architecture. \rev{``SuperPretrain'': ImageNet pre-training of the super network.}}
    \label{tab: effect of PR}
    \begin{threeparttable}
        \begin{tabular}{=c +l +c +c +c +c}
        \toprule
        \multirow{1}*{\textbf{Row}} & \textbf{Method} & \textbf{Total Cost} & \textbf{MAdds} & \textbf{mIOU(\%)}\\
        \midrule
        (1) & Remap $\to$ ArchAdapt $\to$ RemapSuper $\to$ ParamAdapt (\name) & 35.8GHs & 24.17B & \textbf{77.1}  \\
        (2) & Remap $\to$ ArchAdapt $\to$ Remap $\to$ ParamAdapt (FNA~\cite{Fang*2020Fast}) & 35.8GHs & 24.17B & 76.6  \\
        (3) & RandInit $\to$ ArchAdapt $\to$ Remap $\to$ ParamAdapt & 35.8GHs & 24.29B & 76.0  \\
        (4) & Remap $\to$ ArchAdapt $\to$ RandInit $\to$ ParamAdapt & 35.8GHs & 24.17B & 73.0  \\
        (5) & RandInit $\to$ ArchAdapt $\to$ RandInit $\to$ ParamAdapt & 35.8GHs & 24.29B & 72.4  \\
        (6) & Remap $\to$ ArchAdapt $\to$ Pretrain $\to$ ParamAdapt & 547.8GHs & 24.17B & 76.5  \\
        \rowstyle{\color{revcolor}}
        (7) & SuperPretrain $\to$ ArchAdapt $\to$ RemapSuper $\to$ ParamAdapt & 283.8GHs & 24.17B & 72.5 \\
        \rowstyle{\color{revcolor}}
        (8) & SuperPretrain $\to$ ArchAdapt $\to$ Remap $\to$ ParamAdapt & 283.8GHs & 24.17B & 75.9 \\
        \bottomrule
    \end{tabular}
    \end{threeparttable}
\end{table*}

As shown in Tab.~\ref{tab: remap_mech_det} and Tab~\ref{tab: remap_mech_seg}, remapping from the super network demonstrates better performance on both object detection framework RetinaNet~\cite{lin2017focal} and semantic segmentation framework DeepLabv3~\cite{deeplab-v3}. However, for SSDLite~\cite{liu2016ssd,sandler2018mobilenetv2}, remapping parameters from the super network achieves the same mAP as that from the seed network. We deduce this is due to the long training schedule of SSDLite, \ie, 60 epochs. We further perform a long training schedule $2\times$ on RetinaNet (24 epochs in MMDetection~\cite{chen2019mmdetection}). The results in Tab.~\ref{tab: remap_mech_det} show performance promotion that remapping from $\gN_{sup}$ can bring over from $\gN_s$ decays from $1.0\%$ to $0.4\%$ with the training schedule set to $2\times$. It indicates that remapping from the super network for parameter adaptation shows more effectiveness in short training scenarios. This conclusion is somewhat similar to that in \cite{DBLP:journals/corr/abs-1811-08883}, which demonstrates longer training schedules from scratch can achieve comparable results with training with a pre-trained model. We compare the training loss and mAP with different remapping mechanisms in Fig.~\ref{fig: loss/map}. Model training with initial parameters remapped from the super network converges much faster than that remapped from the seed network in early epochs and achieves a higher final mAP in short training schedules. Training with the two remapping mechanisms can achieve similar results in long training schedules, \eg, SSDLite training. It is suggested to remap the parameters from the super network when computation resources are constrained.

\subsection{Effectiveness of Parameter Remapping}
\label{subsec: effect PR}
To evaluate the effectiveness of the parameter remapping paradigm in our method, we attempt to optionally remove the parameter remapping process before the two stages, i.e. architecture adaptation and parameter adaptation. The experiments are conducted with the DeepLabv3~\cite{deeplab-v3} semantic segmentation framework on the Cityscapes dataset~\cite{DBLP:conf/cvpr/CordtsORREBFRS16}. 

Tab.~\ref{tab: effect of PR} shows the complete experiments we perform on parameter remapping. Row (1) denotes the procedure of \name\ and Row (2) denotes the former version which remaps the seed parameters for parameter adaptation. In Row (3) we remove the parameter remapping process before architecture adaptation. In other word, the search is performed from scratch without using the pre-trained network. The mIOU in Row (3) drops by 0.6\% compared to Row (2). Then we remove the parameter remapping before parameter adaptation in Row (4), i.e. training the target architecture from scratch on the target task. The mIOU decreases by 3.6\% compared with (2). When we remove the parameter remapping before both stages in Row (5), it gets the worst performance. In Row (6), we first pre-train the searched architecture on ImageNet and then fine-tune it on the target task. It is worth noting that FNA achieves a higher mIOU by a narrow margin (0.1\%) than the ImageNet pre-trained one in Row (6). We conjecture that this may benefit from the regularization effect of parameter remapping before the parameter adaptation stage. 

\rev{We further pre-train the super network on ImageNet for 100 epochs before architecture adaptation. As shown in Row (7), if we remap the parameters from the super network for latter parameter adaptation, it achieves a low mIOU, 72.5\% with 24.17B MAdds, only a little better than Row (5), which uses random initialization for both architecture and parameter adaptation. This indicates the above super network pre-training on ImageNet contributes little to the whole adaptation process, while it takes a huge cost, 283.8 GPU hours in total, 7.9$\times$ of \name. Because the super network involves a large number of possible paths, \ie candidate architectures, fully pre-training every architecture requires a very long schedule. A regularly scheduled pre-training only brings the super network a poor performance on ImageNet, which helps the latter adaptation very little. 
We further remap the parameters for parameter adaptation from the seed MobileNetV2  in Row (8) instead of the super network in Row (7). It promotes mIOU from 72.5\% in Row (7) to 75.9\%, yet close to Row (3), which searches for the architecture with random initialization. This means that super network pre-training contributes little to both architecture and parameter adaptation and verifies the effectiveness of parameter remapping.}

All the experiments are conducted using the same searching and training settings for fair comparisons. With parameter remapping applied on both stages, the adaptation achieves the best results in Tab.~\ref{tab: effect of PR}. Especially, the remapping process before parameter adaptation tends to provide greater performance gains than the remapping before architecture adaptation. All the experimental results demonstrate the importance and effectiveness of the proposed parameter remapping scheme.

\begin{table}[thbp]
    \centering
    \caption{\rev{Comparing with Net2Net remapping strategies. The experiments are performed with RetinaNet on MS-COCO. The ``Width'', ``Depth'' and ``Kernel'' column indicate the remapping strategy on the width, depth and kernel level respectively. Row (1) represents the seed network MobileNetV2.}}
    \label{tab: comp_n2n}
    \begin{threeparttable}
        \begin{tabular}{+c +c +c +c +c +c}
        \toprule
        \rowstyle{\color{revcolor}}
        \multirow{1}*{\textbf{Row}} & \textbf{Width} & \textbf{Depth} & \textbf{Kernel} & \textbf{MAdds} & \textbf{mAP(\%)}\\
        \midrule
        (1) & - & - & - & 133.05B & 32.8 \\
        (2) & \name & \name & \name & 133.39B & \textbf{33.9} \\
        (3) & Net2Net & \name & \name & 133.39B & 33.7 \\
        (4) & \name & Net2Net & \name & 133.39B & 33.4 \\
        (5) & Net2Net & Net2Net & \name & 133.39B & 33.1 \\
        (6) & Net2Net & Net2Net & Net2Net & 133.39B & N/A \\
        \bottomrule
    \end{tabular}
    \end{threeparttable}
\end{table}

\rev{\subsection{Comparing with Net2Net}
\label{subsec: comp_net2net}
Net2Net~\cite{chen2015net2net} proposes the function-preserving transformations to remap the parameters of one network to a larger one. However, the Net2Net remapping mechanism only involves the expansion cases, \ie Net2WiderNet and Net2DeeperNet, which is inapplicable to most NAS scenarios, while \name\ can be applied on various dimensions, including the width, depth and kernel level with both expansion and shrinkage cases. To evaluate the remapping effectiveness in the expansion case, we perform a series of experiments comparing with Net2Net as shown in Tab.~\ref{tab: comp_n2n}. We expand the widths of the architecture adapted with MobileNetV2 as the seed to introduce the width expansion case for remapping\footnote{\rev{We increase the MBCon expansion ratios from 6 to 8 in several layers. The maximum expansion ratio in MobileNetV2 is 6. The depth increase cases already exist in this adapted architecture, so we do not increase the depth repeatedly.}}. Then we remap the MobileNetV2 parameters to the expanded architecture and train it with RetinaNet on MS-COCO. Row (2) is the result with the \name\ remapping strategy applied on both the width and depth level. When we change the remapping strategy into the Net2Net manner on the width level in Row (3), the mAP decreases by 0.2\%; and on the depth level in Row (4), the mAP decreases by 0.5\%. When the Net2Net remapping strategy is applied on both the width and depth level in Row (5), the network achieves the worst result, 33.1\% mAP, 0.8\% worse than the \name\ based one in Row (2). Noting that all the experiments are equipped with \name\ remapping on the kernel level, Net2Net is not applicable to the kernel-level remapping as shown in Row (6).}

\begin{table*}[htbp]
    \centering
    \caption{Study the strategies of parameter remapping with DeepLabv3 on Cityscapes. ``Wdith-BN" denotes remapping with BN statistics on the width-level. ``Width-Std" and ``Width-L1" denote remapping with std- and $\normlone$ norm- based weight importance on the width-level. ``Kernel-Dilate" denotes remapping with a dilation manner on the kernel-level.}
    \label{tab: methods of PR}
    \begin{threeparttable}
        \begin{tabular}{l c c c c c c}
        \toprule
        \textbf{Method} & Width-BN & Width-Std & Width-L1 & Kernel-Dilate & FNA & \name \\
        \midrule
        \textbf{mIOU(\%)} & 75.8 & 75.8 & 75.3 & 75.6 & 76.6 & \textbf{77.1} \\
        \bottomrule
    \end{tabular}
    \end{threeparttable}
\end{table*}

\rev{We explain the advantages of \name\ remapping strategies over Net2Net as follows. Function-preserving is of critical importance for parameter remapping. 
The remapping strategy in Net2Net follows the preserving principle, but is only available to very limited scenarios. In addition, the width remapping in Net2Net involves the random selection and replication factor computing, which cause the coupling effect between layers and are too complicated to implement. The width remapping in \name\ is much easier for implementation as it is independent in different layers. It also conforms to the function preserving principle as analyzed in Appendix~\ref{subsec: func_pres}, but is more flexible and better performing than Net2Net.}

\rev{In Net2Net~\cite{chen2015net2net}, the identity matrix is introduced for deeper network remapping to keep the function-preserving principle. However, in popular ResNet or MobileNet, the identity residual connection is used in one convolution block, which already holds the function-preserving property. To still keep the preserving principle, the weights in the convolution block should be zero-initialized\footnote{Slight noises are added to enable updating.}. This manner yet introduces too many new weights to learn from scratch. \name\ reuses the weights from former layers and achieves better performance, which also indicates the preserving principle is not so necessary on the depth level. Moreover, the reusability of parameters from different depths is also verified in previous studies~\cite{veit2016residual,greff2016highway} by indicating reordering the building blocks in a residual network has a modest impact on performance.}

\begin{table*}[h!]
    \centering
    \caption{Results of random search experiments with the RetinaNet framework on MS-COCO. ``DiffSearch": differentiable NAS. ``RandSearch": random search. \rev{``RandSample'': random sampling.} The other abbreviation definitions are the same as Tab.~\ref{tab: effect of PR}.}
    \label{tab: randsearch}
    \begin{threeparttable}
        \begin{tabular}{=c +l +c +c +c}
        \toprule
        \multirow{1}*{\textbf{Row}} & {\textbf{Method}} & \textbf{Total Cost} & \textbf{MAdds} & \textbf{mAP(\%)} \\
        \midrule
        (1) & DetNAS~\cite{DBLP:journals/corr/abs-1903-10979} & 68 GDs & 133.26B & 33.3    \\
        (2) & Remap $\to$ DiffSearch $\to$ Remap $\to$ ParamAdapt & 9.2 GDs & 133.03B & \textbf{33.9}   \\
        (3) & Remap $\to$ RandSearch $\to$ Remap $\to$ ParamAdapt & 9.9 GDs & 133.11B & 33.5    \\
        (4) & RandInit $\to$ RandSearch $\to$ Remap $\to$ ParamAdapt & 9.9 GDs & 133.08B & 31.5   \\
        (5) & Remap $\to$ RandSearch $\to$ RandInit $\to$ ParamAdapt & 9.9 GDs & 133.11B & 25.3    \\
        (6) & RandInit $\to$ RandSearch $\to$ RandInit $\to$ ParamAdapt & 9.9 GDs & 133.08B & 24.9    \\
        \rowstyle{\color{revcolor}}
        (7) & Remap $\to$ RandSample $\to$ Remap $\to$ ParamAdapt & - & 133.17$_{\pm 0.17}$B & 32.6$_{\pm 0.54}$ \\
        \bottomrule
    \end{tabular}
    \end{threeparttable}
\end{table*}

\subsection{Studying Parameter Remapping Strategies}
\label{subsec: studyPR}
We explore more strategies for the parameter remapping paradigm. All the experiments are conducted with the DeepLabv3~\cite{deeplab-v3} framework on the Cityscapes dataset~\cite{DBLP:conf/cvpr/CordtsORREBFRS16}. We make exploration from the following respects. For simplicity, we denote the weights of the seed network and the new network on the remapping dimension (output/input channel) as $\mW_s = (\mW_s^{(1)} \dots \mW_s^{(p)})$ and $\mW_n = (\mW_n^{(1)} \dots \mW_n^{(q)})$.

\subsubsection{Remapping with BN Statistics on Width-level}
We review the formulation of batch normalization~\cite{DBLP:conf/icml/IoffeS15} as follows, 
\begin{equation}
    y_i \gets \gamma \frac{x_i - \mu_{\mathcal{B}}}{\sqrt{\sigma^2_{\mathcal{B}}+\epsilon}} + \beta\text{,}
\end{equation}
where $x_i = (x_i^{(1)} \dots x_i^{(p)})$ denotes the $p$-dimensional input tensor of the $i$th layer, $\gamma \in \R^p$ denotes the learnable parameter which scales the normalized data on the channel dimension. We compute the absolute values of $\gamma$ as $|\gamma| = (|\gamma^{(1)}| \dots |\gamma^{(p)}|)$. When remapping the parameters on the width-level, we sort the values of $|\gamma|$ and map the parameters with the sorted top-$q$ indices. More specifically, we define a weights remapping function in Algo.~\ref{algo: mapping}, where the reference vector $\vv$ is $|\gamma|$.

\begin{algorithm}[ht]
  \label{algo: mapping}
  \caption{Weights Remapping Function}
  \LinesNumbered
  \KwIn{the seed weights $\mW_s$ and the new \\
  weights $\mW_n$, the reference vector $\vv$}
  // get indices of topk values of the vector\\
  $\va \gets topk$-$indices(\vv, k=q)$\\
  // sort the indices\\
  $sort(\va)$\\
  \For{$i \in 1, 2, \dots, q$}{
      $\mW^{(i)}_n = \mW^{(\va[i])}_s$
  }
  \KwOut{$\mW_n$ with remapped values}
\end{algorithm}

\begin{figure}[thbp]
  \centering
  \includegraphics[width=0.45\columnwidth]{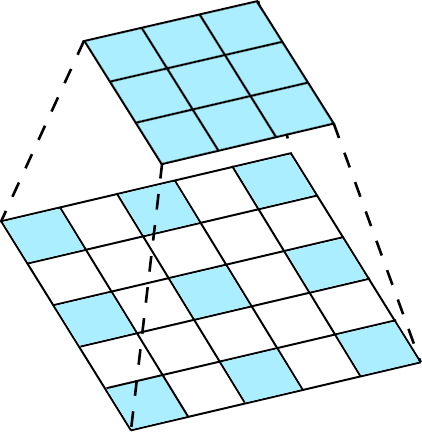}
  \captionof{figure}{Parameter Remapping on the kernel-level with a dilation manner.}
  \label{fig: kernel_dilate}
\end{figure}

\subsubsection{Remapping with Weight Importance on Width-level}
We attempt to use a canonical form of convolution weights to measure the importance of parameters. Then we remap the seed network parameters with great importance to the new network. The remapping operation is conducted based on Algo.~\ref{algo: mapping} as well. We experiment with two canonical forms of weights to compute the reference vector, the standard deviation of $\mW_s$ as $(std(\mW_s^{(1)}) \dots std(\mW_s^{(p)}))$ and the $\normlone$ norm of $\mW_s$ as $(|| \mW_s^{(1)} ||_1 \dots || \mW_s^{(p)} ||_1)$.

\subsubsection{Remapping with Dilation on Kernel-level}
We experiment with another strategy of parameter remapping on the kernel-level. Different from the method defined in Sec.~\ref{subsec: params_remap}, we remap the parameters with a dilation manner as shown in Fig.~\ref{fig: kernel_dilate}. The values in the convolution kernel without remapping are all assigned with $0$. It is formulated as
\begin{equation}
    \mW^{k \times k}_{:,:,h,w} = \begin{cases}
        \mW^{3 \times 3}_{:,:,h,w} & \text{if } h, w = 1 + i \cdot \frac{k-1}{2} \text{ and } i=0,1,2\\
        0 & \text{otherwise}
    \end{cases}\text{,}
    \label{eq: kernel_dilate}
\end{equation}
where $\mW^{k \times k}$ and $\mW^{3 \times 3}$ denote the weights of the new network and the seed network respectively, $h, w$ denote the spatial indices.

Tab.~\ref{tab: methods of PR} shows the experimental results and all the searched models hold the similar MAdds. The network adaptation with the parameter remapping paradigm defined in Sec.~\ref{subsec: params_remap} achieves the best results. Furthermore, the remapping operation of \name\ is easier to implement compared to the several aforementioned ones. We explore limited number of methods to implement the parameter remapping paradigm. How to conduct the remapping strategy more efficiently remains a significative work.

\subsection{Random Search \rev{\& Sampling }Experiments}
\label{subsec: randsearch}
We perform the Random Search (RandSearch) experiments with the RetinaNet~\cite{lin2017focal} framework on the MS-COCO~\cite{COCO} dataset. All the results are shown in the Tab.~\ref{tab: randsearch}. In Row (3), we purely replace the original differentiable NAS (DiffSearch) method in \name\ with the random search method. \rev{The random search is performed similarly to that indicated by \cite{stamoulis2019single,li2020random,Yu2020Evaluating}, and takes the same computation cost as the search in \name\ for fair comparisons. Specifically, we randomly sample 24 architectures (to keep the same search cost) from the search space. All the samples are guaranteed to have similar MAdds with the model searched by FNA. Then we train every sampled model for 1 epoch and select the one with the highest mAP to train it completely.}

We observe that \name\ with RandSearch achieves comparable results with our original method. It further confirms that \name\ is a general framework for network adaptation and has great generalization ability. NAS is only an implementation tool for architecture adaptation. The whole framework of \name\ can be treated as a NAS-method agnostic mechanism. It is worth noting that even using random search, our \name\ still outperforms DetNAS~\cite{DBLP:journals/corr/abs-1903-10979} with 0.2\% mAP better and 150M MAdds fewer. We further conduct similar ablation studies with experiments in Sec.~\ref{subsec: effect PR} about the parameter remapping scheme in Row (4) - (6). All the experiments further support the effectiveness of the parameter remapping scheme.

\rev{To evaluate the specific improvements brought by architecture search, we perform a random sampling~\cite{Yang2020NAS} experiment in Row (7) with parameter remapping applied as well. We randomly sample 5 architectures from the search space with similar MAdds with the FNA model. Then we train all the sampled model completely with parameter remapping applied. The mean MAdds and mAP of the sampled models are reported in Row (7). The average (randomly sampled) architecture achieves a 1.3\% worse mAP with slightly larger MAdds.}

{\color{revcolor}
% \section{Analysis and Discussion}
% \label{sec: analy}
% In this section, we first give more insights into the working mechanism and advantages of the network adaptation. Then we demonstrate that both the width- and kernel- level remapping conforms to the function-preserving principle.

% \begin{figure*}[!h]
%     \centering
%     \subfigure[adaptation w/ seed]
%     {\begin{minipage}[b]{0.44\textwidth}
%         \includegraphics[width=\textwidth]{figs_r1/arch-trensfer.pdf}
%         \label{fig: adapt_w_seed}
%     \end{minipage}}
%     \hspace{5pt}
%     \subfigure[adaptation w/o seed]
%     {\begin{minipage}[b]{0.29\textwidth}
%         \includegraphics[width=\textwidth]{figs_r1/super-pretrain.pdf}
%         \label{fig: adapt_wo_seed}
%     \end{minipage}}
%     \caption{\rev{Comparisons between the adaptation processes with and without the seed. (a) shows the adaptation process with a seed, where the search is an architecture-level fine-tuning and is performed near the seed. (b) shows the adaptation without a seed, where the starting point of the search is randomly selected, and it is hard to find the solution in the new task.}}
% \end{figure*}
\begin{figure*}[!h]
    \centering
    \includegraphics[width=0.8\textwidth]{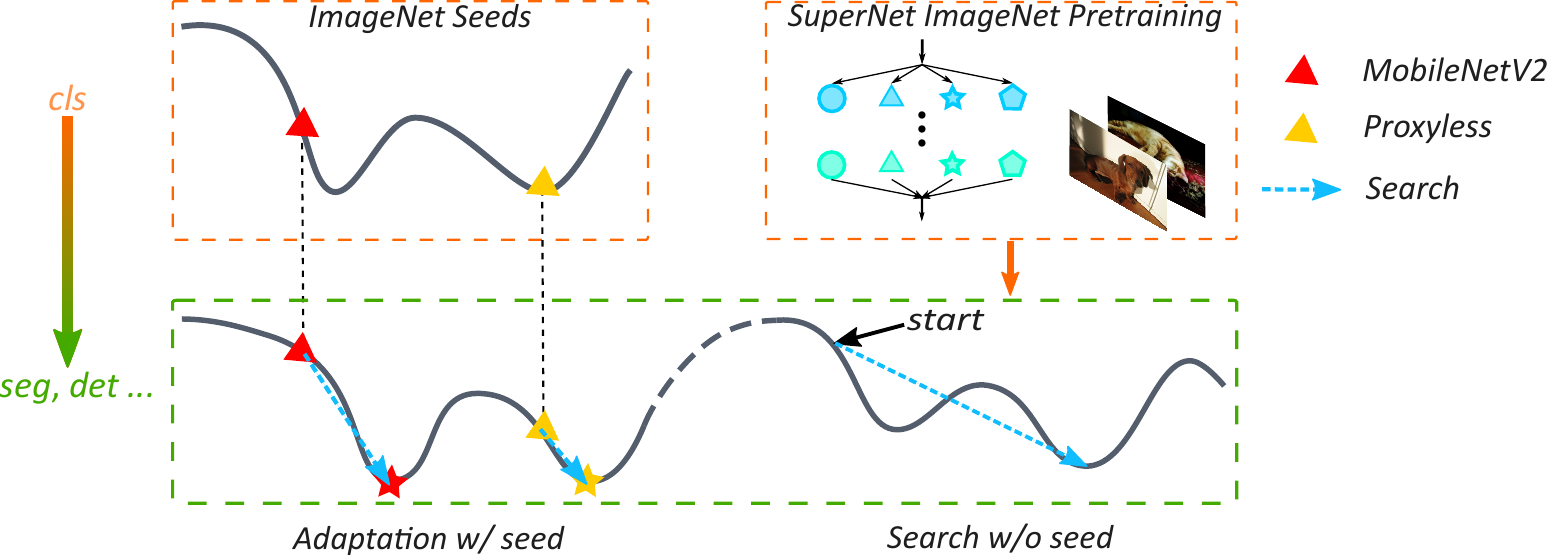}
    \label{fig: adapt_comp}
    \caption{\rev{Comparisons between the architecture adaptation processes with the seed (left) and search without the seed (right). 
    We use the curve to represent the architecture space, which is supposed to be a high-dimensional space, but we plot the 2-D curve for simplicity. Each point in the curve represents an architecture, and the minimum point is a solution for adaptation.
    \emph{Left} shows the adaptation process with a seed, where the search is a architecture-level fine-tuning and is performed near the seed. \emph{Right} shows the search without a seed, where the starting point of the search is randomly selected, and it is hard to find the solution in the new task because of the large search range.}}
\end{figure*}

\section{Working Mechanism Understanding}
\label{sec: work_mech}
We first give more insights into the working mechanism and advantages of the network adaptation as follows. The working mechanism of \name\ is supported by the information reuse of the seed network, which relies on the proposed parameter remapping scheme. The advantages of adaptation with the seed over that without a seed mainly include two aspects, \ie parameter and architecture. \name\ provides both parameter- and architecture- level transfer learning, \ie, fine-tuning. Parameter-level fine-tuning is widely applied in deep learning but not applied in NAS, while the concept of architecture-level fine-tuning between tasks is of critical importance in the NAS field.

For parameters, the seed network can directly provide effective parameters by remapping for both search and the target architecture retraining. However, adaptation without a seed normally requires the cumbersome super network pre-training. And as shown in Tab.~\ref{tab: effect of PR}, even if the costly super network pre-training is performed, the parameters from the super network still cannot support the parameter adaptation, as the super network involves too many possible architectures and the regular training schedule is far not enough. 

We illustrate the advantages from the architecture aspect as follows. It is revealed in many previous NAS works~\cite{MnasNet,fbnet,cai2018proxylessnas,guo2020single} that there exist non-unique solutions with different architecture shapes but similar high performance in the same architecture design space. For example, SPOS~\cite{guo2020single} achieves a 74.7\% Top-1 ImageNet accuracy with 328M MAdds while Proxyless (mobile) achieves a 74.6\% accuracy with 320M MAdds, which hold quite different architectures. As shown in the upper left of Fig.~\ref{fig: adapt_comp}, the handcrafted network MobileNetV2 is not located in the solution of the classification task. The Proxyless~\cite{cai2018proxylessnas} network is adapted to a non-unique solution by the NAS method. When transferred to a new task, \eg semantic segmentation or object detection, both architectures deviate from the solutions of the new task due to differences between tasks. However, because of the common characteristics between tasks, \eg requiring for semantic information extraction, the architectures do not deviate too much. The architecture adaptation is a fine-tuning process on the architecture level. As the parameters of the super network are remapped from the seed, the parameters of architectures near the seed dominate the others. Therefore, the search is performed near the seed. In addition, the seed Proxyless shows better performance on object detection than MobileNetV2 before adaptation, and the adaptation makes fewer changes to Proxyless than MobileNetV2 as shown in Fig.~\ref{fig: fna_networks} and Fig.~\ref{fig: nas_networks} of Appendix. This indicates Proxyless is closer to the non-unique solution than MobileNetV2. The differences between tasks make the adaptation necessary, while the similarities between tasks make the architecture-level fine-tuning so effective.

On the contrary, search without the seed usually requires a cumbersome super network pre-training, which uniformly samples the architectures and trains them equally. This means the parameters of all the architectures are at the same starting line for search on the target task. As shown in the right of Fig.~\ref{fig: adapt_comp}, the starting point of the search is randomly selected, \eg it depends on the first stochastic gradient descent of differentiable NAS or the randomly initialized population in EA-based NAS. The search range without the seed is far larger than that with the seed, so it is more difficult for the search algorithm to find the solution. The experimental results in Tab.~\ref{tab: effect of PR} also show that searching without the seed performs worse than that with the seed.

}

% --------------------------------Section Partition------------------------------- %
\section{Conclusion}
\label{sec: Conclusion}
In this paper, we propose a fast neural network adaptation method (\name) with a novel parameter remapping paradigm and the architecture search method. We adapt the manually designed network MobileNetV2 to semantic segmentation, object detection and human pose estimation tasks on both architecture- and parameter- level. The generalization ability of \name\ is further demonstrated on both ResNets and NAS networks. The parameter remapping paradigm takes full advantages of the seed network parameters, which greatly accelerates both the architecture search and parameter fine-tuning process. With our \name\ method, researchers and engineers could fast adapt more pre-trained networks to various frameworks on different tasks. As there are lots of ImageNet pre-trained models available in the community, we could conduct adaptation with low cost and do more applications, e.g., face recognition, depth estimation, etc. Towards real scenarios with dynamic dataset or task demands, \name\ is a good solution to adapt or update the network with negligible cost. For researchers with constrained computation resources, \name\ can be an efficient tool to perform various explorations on computation consuming tasks.

\section*{Acknowledgements}
This work was in part supported by NSFC (No. 61876212 and No. 61733007), Zhejiang Lab (No. 2019NB0AB02) and HUST-Horizon Computer Vision Research Center. We thank Liangchen Song, Wenqiang Zhang, Yingqing Rao and Jiapei Feng for the discussion and assistance.

\bibliographystyle{IEEEtran}
\bibliography{fna}

\if\usecolor1
{\color{revcolor}
\appendices
\section{}
\subsection{Function Preserving Demonstration}
\label{subsec: func_pres}
In this section, we demonstrate that the \name\ remapping strategy conforms to the function-preserving principle on the width and kernel expansion level.
\subsubsection{Width-level Preserving}
As typical networks, \eg ResNet~\cite{he2016deep}, MobileNet~\cite{howard2017mobilenets,sandler2018mobilenetv2}, construct the architecture with convolution blocks, \ie a set of several convolution layers, we analyze the width-level function preserving principle within a convolution block. We take the MobileNetV2 block (MBConv) as example. For simplicity, we assume the input tensor as a vector form $\vx \in \R^{p}$ and all the convolution kernels in MBConv are $1 \times 1$ ones. Normally, an MBConv can be formulated as,
\begin{equation}
    \vo = [f_2 \circ d \circ f_1](\vx)\text{,}
\end{equation}
where $\vo \in \R^{q}$ is the output tensor, $f_1$ and $f_2$ are the two linear transformations whose weights are denoted as $\mW_1 \in \R^{tp \times p}$ and $\mW_2 \in \R^{q \times tp}$ respectively, $d$ is the middle depth-wise convolution whose weights are denoted as $\mW_d \in \R^{tp}$, and $t$ is the width expansion ratio. $\vo$ is computed as, 
\begin{equation}
\begin{aligned}
    \vo_1 = f_1(\vx) &= \mW_1 \cdot \vx^T\\
    \vo_2 = d(\vo_1) &= \mW_d^T \odot \vo_1\\
    \vo = f_2(\vo_2) &= \mW_2 \cdot \vo_2 \text{,}
\end{aligned}
\end{equation}
where $\odot$ denotes the depth-wise convolution, \ie element-wise multiplication for the supposed vector-input case.

For the width expanded MBConv, we assume the expansion width is $tp+r$, where $r$ is the increased channel numbers. The expanded MBConv is formulated as,
\begin{equation}
    \vo' = [f_2' \circ d' \circ f_1'](\vx)\text{,}
\end{equation}
where $\vo' \in \R^{q}$ is the output tensor, $f_1'$ and $f_2'$ have weights of $\mW_1' \in \R^{(tp+r) \times p}$ and $\mW_2' \in \R^{q \times (tp+r)}$ respectively, $d'$ have the weights of $\mW_d' \in \R^{(tp+r)}$. With the \name\ parameter remapping applied, the weights of the expanded MBConv have the following properties.
\begin{equation}
    \mW_1' = \begin{bmatrix}
        \mW_1 \\ \bm{0}_{r, p}
    \end{bmatrix}\text{, }
    \mW_d = [\mW_d \quad \bm{0}_{1, r}]\text{, }
    \mW_2' = [\mW_2 \quad \bm{0}_{q, r}]\text{.}\\
\end{equation}
Then $\vo'$ can be computed as,
\begin{equation}
\begin{aligned}
    \vo_1' = f_1'(\vx) &= \mW_1' \cdot \vx^T 
                    = \begin{bmatrix}
                        \mW_ 1 \cdot \vx^T \\ \bm{0}_{r, 1}
                    \end{bmatrix}
                    =\begin{bmatrix}
                        \vo_1 \\ \bm{0}_{r, 1}
                    \end{bmatrix}\\
    \vo_2' = d'(\vo_1') &= \mW_d'^T \odot \vo_1'
                        = \begin{bmatrix}
                            \mW_d^T \odot \vo_1 \\ \bm{0}_{r, 1}
                        \end{bmatrix}
                        =\begin{bmatrix}
                            \vo_2 \\ \bm{0}_{r, 1}
                        \end{bmatrix}\\
    \vo' = f_2'(\vo_2') &= \mW_2' \cdot \vo_2' 
                    = [\mW_2 \quad \bm{0}_{q, r}] \cdot \begin{bmatrix}
                        \vo_2 \\ \bm{0}_{r, 1}
                    \end{bmatrix}\\
                    &= \mW_2 \cdot \vo_2 = \vo\text{.}
\end{aligned}
\end{equation}
The above process demonstrates that the width-level remapping defined in \name\ also conforms to the function preserving principle.

\subsubsection{Kernel-level Preserving}
We take the input/output feature map with 1 channel, denoted as $\mX \in \R^{H \times W}$ for input and $\mO \in \R^{H \times W}$ for output respectively, and assume $H = W$ for simplicity. Suppose the convolution of the seed network has a kernel size of $k \times k$, whose weights are denoted as $\mW \in \R^{K \times K}$. The weights of the target network are denoted as $\mW' \in \R^{(k+l) \times (k+l)}$, where $l$ is the increased kernel size. Normally, the padding size for convolution is computed as $p = (k - 1) / 2$, where $k$ is the kernel size. We denote the padded feature map tensor in the seed network as $\bar{\mX} \in \R^{(H+k-1) \times (W+k-1)}$, and the padded tensor in the target network as $\bar{\mX}' \in \R^{(H+k+l-1) \times (W+k+l-1)}$. For commonly used zero-padding, we have
\begin{equation}
    \bar{\mX}'[i, j] = \begin{cases}
        \bar{\mX}[i-\frac{l}{2}, j-\frac{l}{2}] & \mbox{if } \frac{l}{2}<i, j \le H+\frac{l}{2}\\
        0 & \mbox{otherwise}
    \end{cases}\text{.}
\end{equation}
With the \name\ remapping strategy, we have
\begin{equation}
    \mW'[i, j] = \begin{cases}
        \mW[i-\frac{l}{2}, j-\frac{l}{2}] & \mbox{if } \frac{l}{2} < i, j \le k+\frac{l}{2}\\
        0 & \mbox{otherwise}
    \end{cases}\text{.}
\end{equation}
The output $\mO$ of the seed network is computed as,
\begin{equation}
    \mO[h, w] = \sum_{i,j=1}^{k} \bar{\mX}[h+i-1, w+j-1]\mW[i, j]\text{,}
\end{equation}
where $1 \le h, w \le H$. Then, the output $\mO' \in \R^{H \times W}$ of the target network can be computed as,
\begin{equation}
\begin{aligned}
    &\mO'[h, w] = \sum_{i,j=1}^{k+l} \bar{\mX}'[h+i-1, w+j-1]\mW'[i, j]\\
    &= \sum_{i,j=\frac{l}{2}+1}^{k+\frac{l}{2}} \bar{\mX}[h+i-1-\frac{l}{2}, w+j-1-\frac{l}{2}]\mW[i-\frac{l}{2}, j-\frac{l}{2}]\\
    &= \sum_{i,j=1}^{k} \bar{\mX}[h+i-1, w+j-1]\mW[i, j]\\
    &= \mO[h, w]
\end{aligned}
\end{equation}
The above computing process demonstrates the function-preserving property of \name\ remapping on the kernel-level.

\begin{figure}[thbp]
    \centering
    \includegraphics[width=\linewidth]{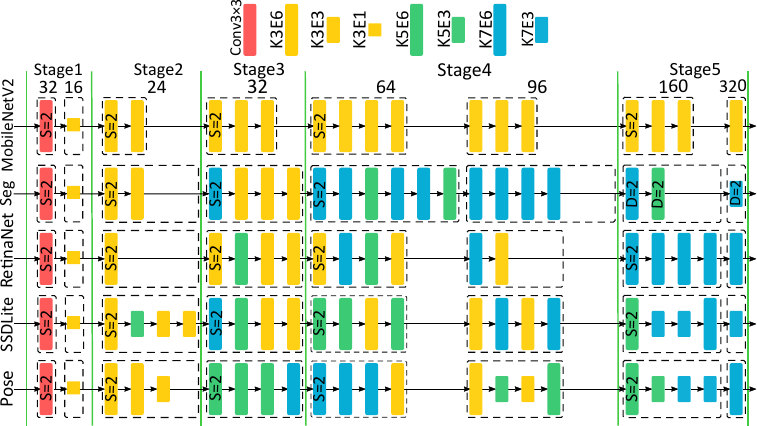}
    \caption{Architectures adapted by \name\ on different tasks and frameworks \rev{(better viewed in colored PDF)}. Each MBConv block is denoted as a colored rectangle. Different colors and shapes of rectangles represent different settings of convolution blocks. ``Conv$3\times3$" denotes the conventional $3\times3$ convolution followed by a batch normalization layer and a ReLU layer (commonly ReLU6 in MobileNetV2-based networks). ``KxEy" denotes the MBConv with ``x" kernel size and ``y" expansion ratio. ``dila" denotes the dilation ratio of the convolution, which is commonly used in the last stage of semantic segmentation networks~\cite{yu2015multi,deeplab-v3}. ``Seg'' and ``Pose'' denote the architectures for semantic segmentation and human pose estimation respectively. \rev{``S=2'' and ``D=2'' denote the stride is 2 and the dilation ratio is 2 respectively. MBConv blocks with the same output channel numbers are contained in a dashed box.}}
    \label{fig: fna_networks}
\end{figure}

\begin{figure}[!t]
    \centering
    \includegraphics[width=\linewidth]{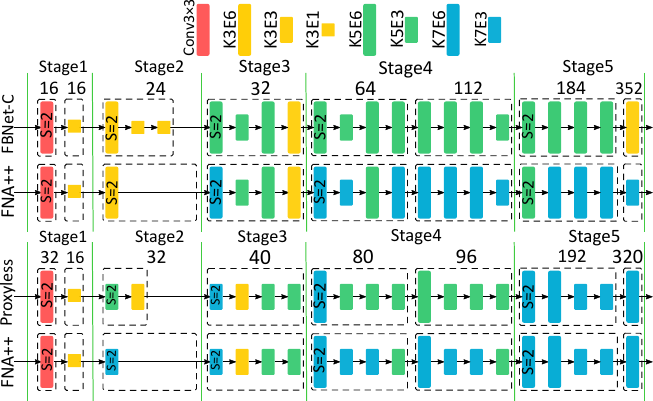}
    \caption{Architectures of NAS networks and ones adapted by \name\ \rev{(better viewed in colored PDF)}. All the abbreviations and definitions are the same as that in Fig.~\ref{fig: fna_networks}.}
    \label{fig: nas_networks}
\end{figure}

\begin{figure}[t!]
    \centering
    \subfigure[training loss]
    {\includegraphics[width=0.8\linewidth]{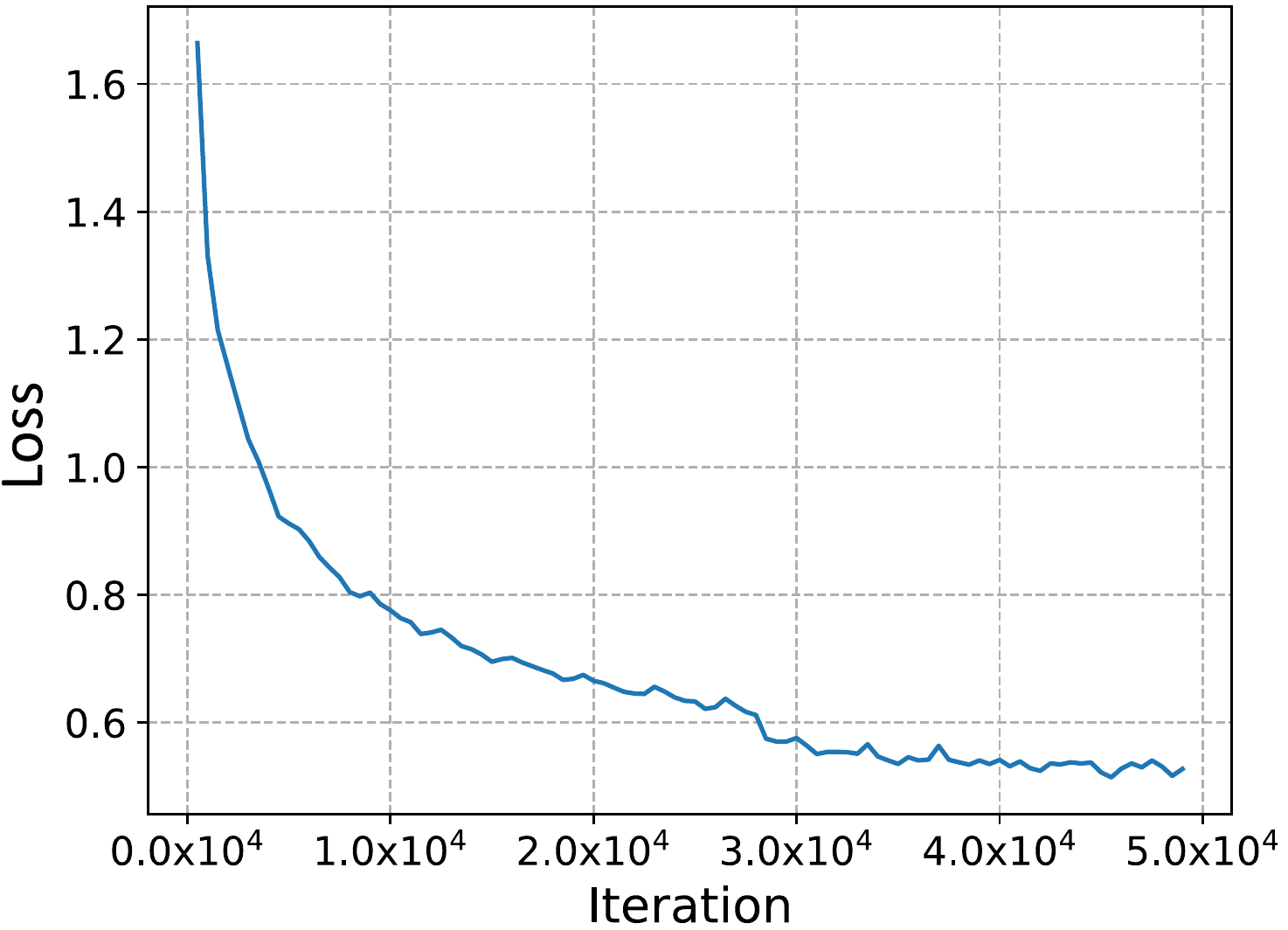}
     \label{fig: search_loss}
    }
    \subfigure[sampled MAdds]
    {\includegraphics[width=0.8\linewidth]{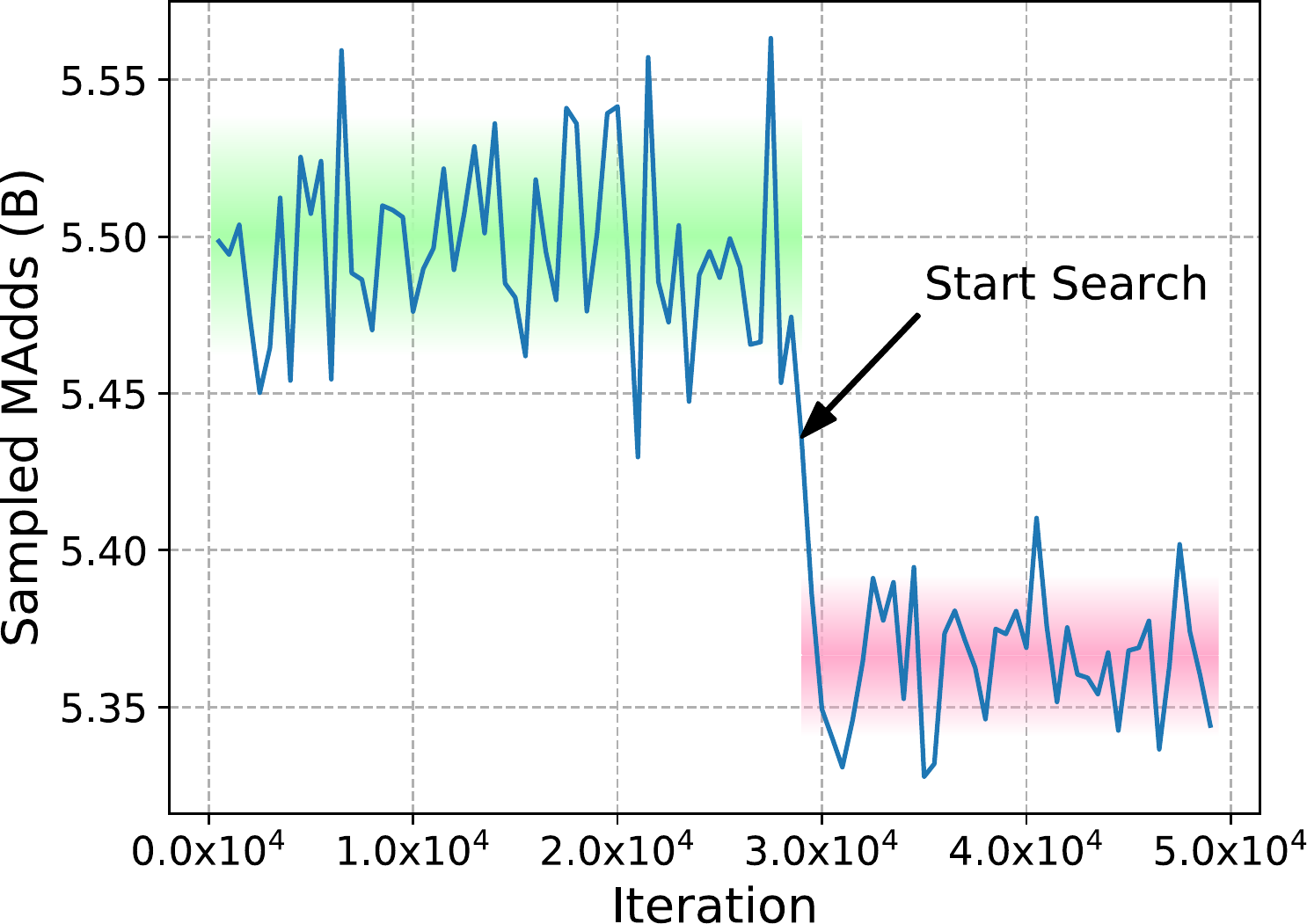}
    \label{fig: madds_curve}
    }
    \caption{\rev{Curves of the training loss (a) and the sampled architecture MAdds (b) during search.}\label{fig: curve}}
\end{figure}

\subsection{Architecture Characteristics for Different Tasks}
\label{subsec: anal_arch}
The architectures adapted to different tasks and frameworks are shown in Fig.~\ref{fig: fna_networks}. We find that the architectures are modified from the seed MobileNetV2 to fit the characteristics of new tasks. We summarize the changes as follows which may be heuristic for future architecture design in the community. Compared with MobileNetV2, the adapted architectures prefer larger-kernel convolution blocks for larger receptive fields. The architecture adaptation reallocates the computation cost of the seed network. Specifically, for semantic segmentation with DeepLabv3~\cite{deeplab-v3}, down-sampling is not performed in the last stage and dilated convolutions are instead used for obtaining larger receptive fields. The layers in the last stage hold huge computation cost due to the high resolution with large channel numbers. After architecture adaptation, the depth and width in stage 5 of the architecture for semantic segmentation are decreased and the kernel sizes in the former stages are enlarged. For object detection, as the input resolution of RetinaNet is much higher than that of the lightweight framework SSDLite, the former layers account for a large percentage of the computation cost. The adapted architecture prefers to contain fewer layers in stage 2 of RetinaNet than SSDLite. For human pose estimation, some width expansion ratios are shrunken in return for kernel sizes enlarging.

\subsection{Search Curves}
We plot the training loss curve of the super network during search on RetinaNet in Fig.~\ref{fig: search_loss}. The training loss converges obviously as the search terminates. And we plot the MAdds of the sampled architectures during the search in Fig.~\ref{fig: madds_curve}. The sampled MAdds fluctuate in a wide range with a large mean value before the architecture parameter updating. With the search starting, the MAdds decrease to a smaller value and the range is narrower, which benefits the optimization of the MAdds during search. These two figures clearly demonstrates the bi-objective optimization process of both accuracy and MAdds.

}

\begin{figure*}[hbp]
    \centering
    \subfigure{
    \begin{minipage}[b]{0.31\linewidth}
        \includegraphics[width=1\linewidth]{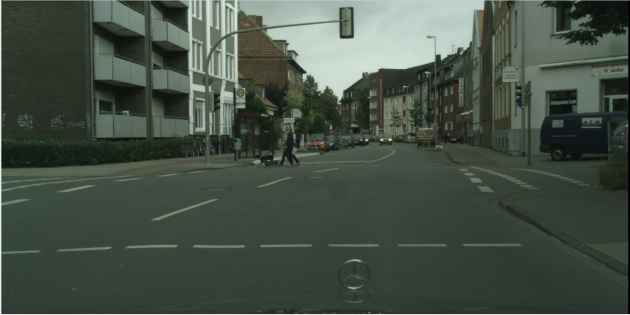}\vspace{4pt}
        \includegraphics[width=1\linewidth]{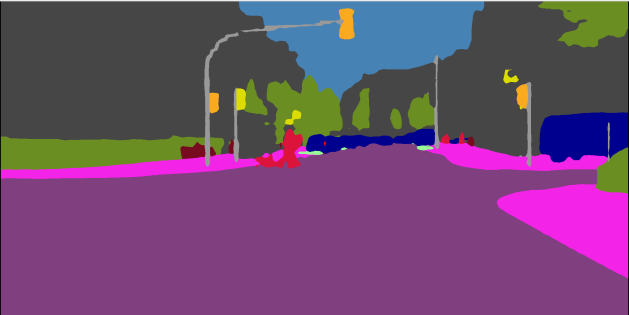}
    \end{minipage}
    }
    \subfigure{
    \begin{minipage}[b]{0.31\linewidth}
        \includegraphics[width=1\linewidth]{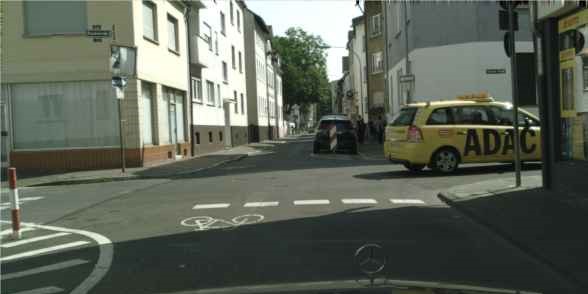}\vspace{4pt}
        \includegraphics[width=1\linewidth]{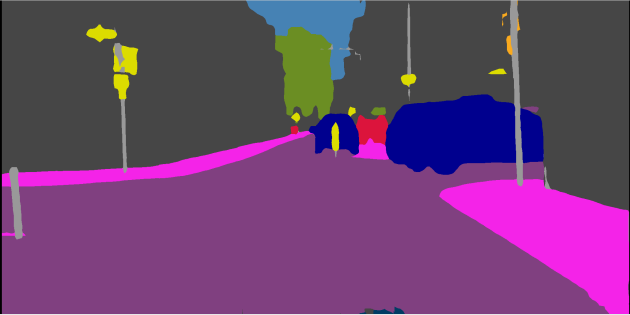}
    \end{minipage}
    }
    \subfigure{
    \begin{minipage}[b]{0.31\linewidth}
        \includegraphics[width=1\linewidth]{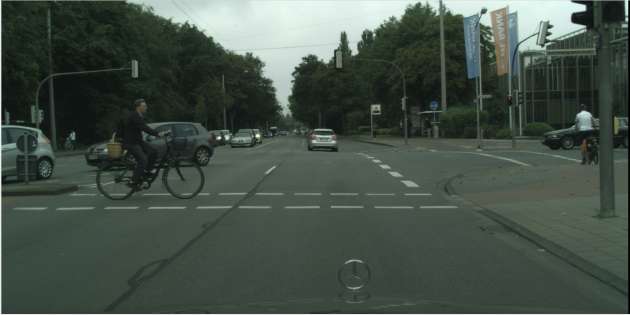}\vspace{4pt}
        \includegraphics[width=1\linewidth]{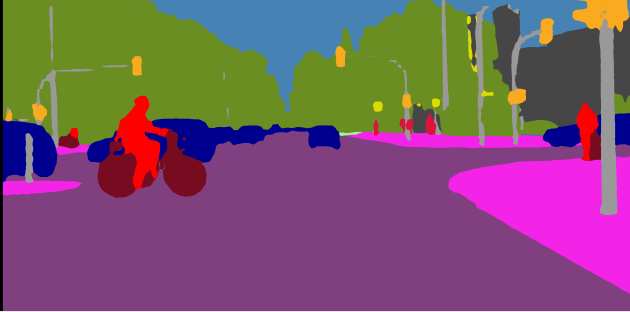}
    \end{minipage}
    }
    \caption{Visualization of semantic segmentation results on the Cityscapes validation dataset. \emph{upper:} the input images. \emph{bottom:} the prediction results.}
\end{figure*}

\begin{figure*}[hbp]
    \centering
    \subfigure{
    \begin{minipage}[b]{0.31\linewidth}
        \includegraphics[width=1\linewidth]{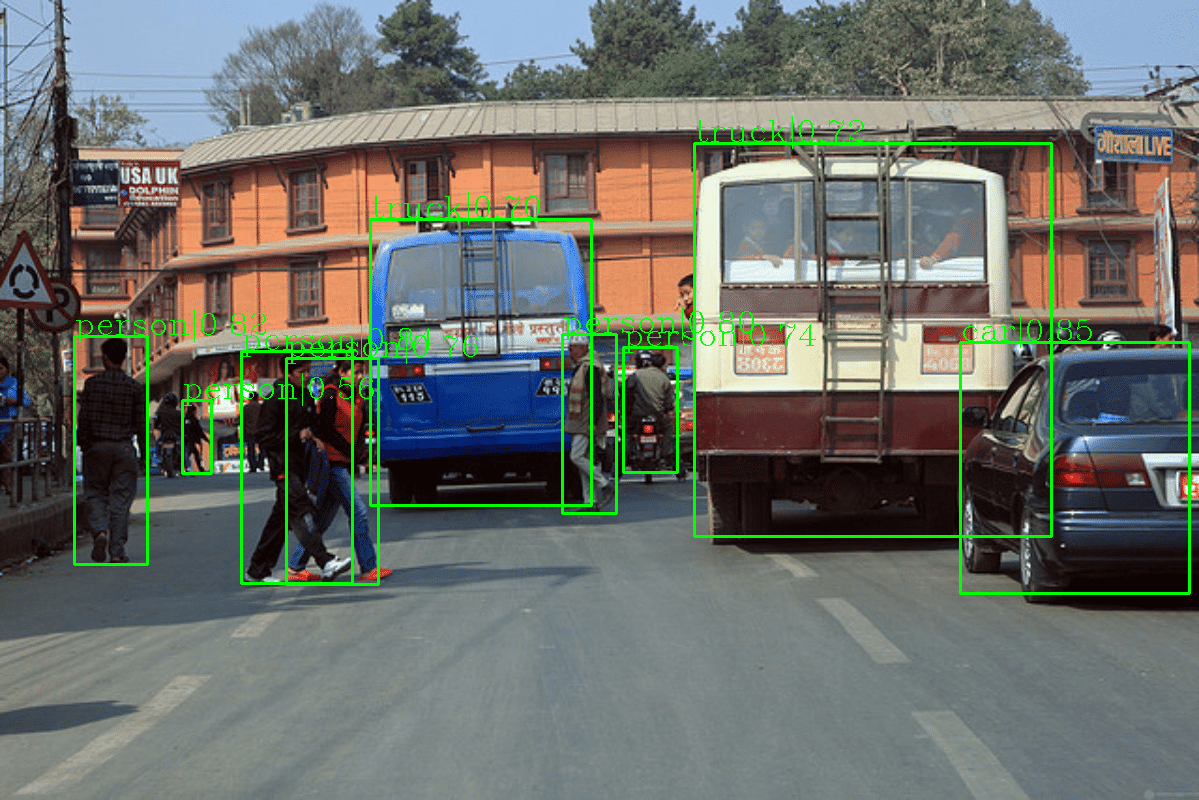}\vspace{4pt}
        \includegraphics[width=1\linewidth]{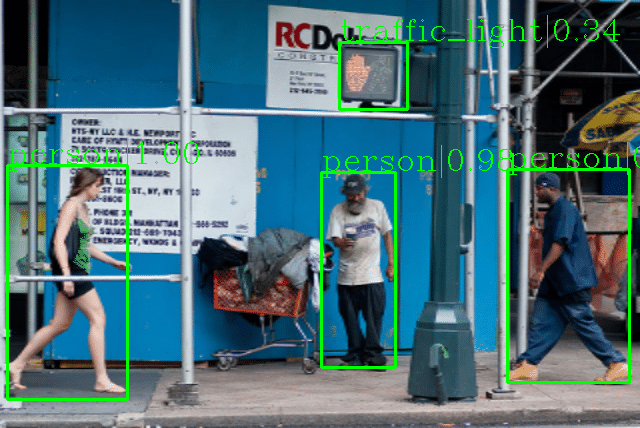}
    \end{minipage}}
    \subfigure{
    \begin{minipage}[b]{0.31\linewidth}
        \includegraphics[width=1\linewidth]{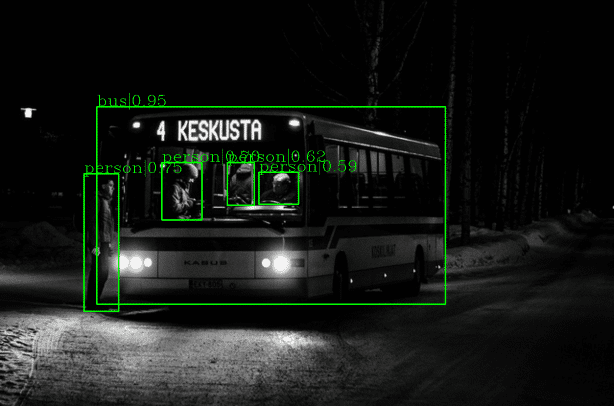}\vspace{4pt}
        \includegraphics[width=1\linewidth]{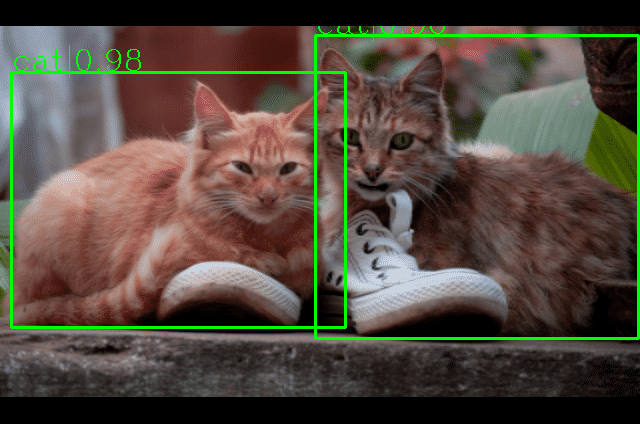}
    \end{minipage}}
    \subfigure{
    \begin{minipage}[b]{0.31\linewidth}
        \includegraphics[width=1\linewidth]{figs/vis/det/RetinaNet_3.png}\vspace{4pt}
        \includegraphics[width=1\linewidth]{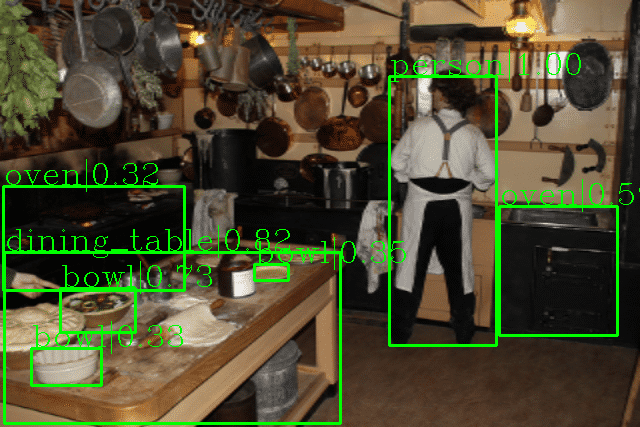}
    \end{minipage}}
    \caption{Visualization of object detection results on the MS-COCO validation dataset. \emph{upper:} results of RetinaNet. \emph{bottom:} results of SSDLite.}
\end{figure*}

\begin{figure*}[hbp]
    \centering
    \subfigure{
    \begin{minipage}[b]{0.31\linewidth}
        \includegraphics[width=1\linewidth]{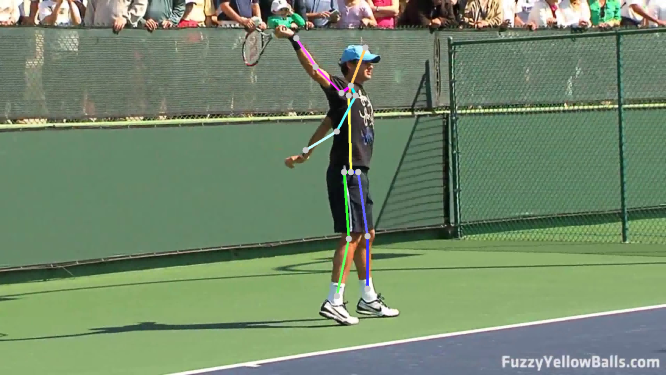}\vspace{4pt}
        \includegraphics[width=1\linewidth]{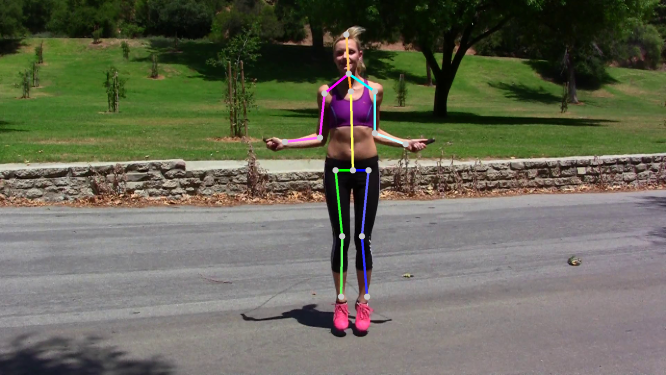}
    \end{minipage}}
    \subfigure{
    \begin{minipage}[b]{0.31\linewidth}
        \includegraphics[width=1\linewidth]{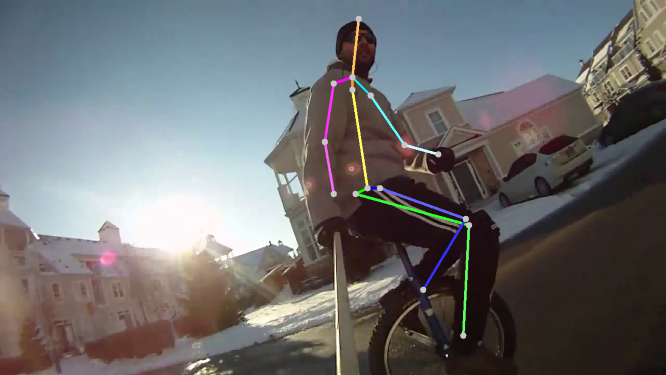}\vspace{4pt}
        \includegraphics[width=1\linewidth]{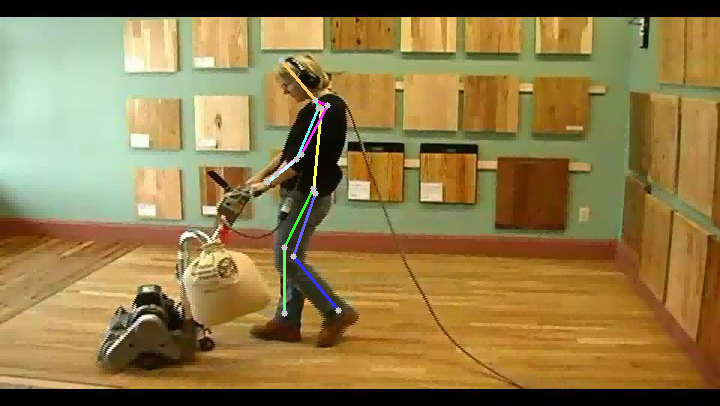}
    \end{minipage}}
    \subfigure{
    \begin{minipage}[b]{0.31\linewidth}
        \includegraphics[width=1\linewidth]{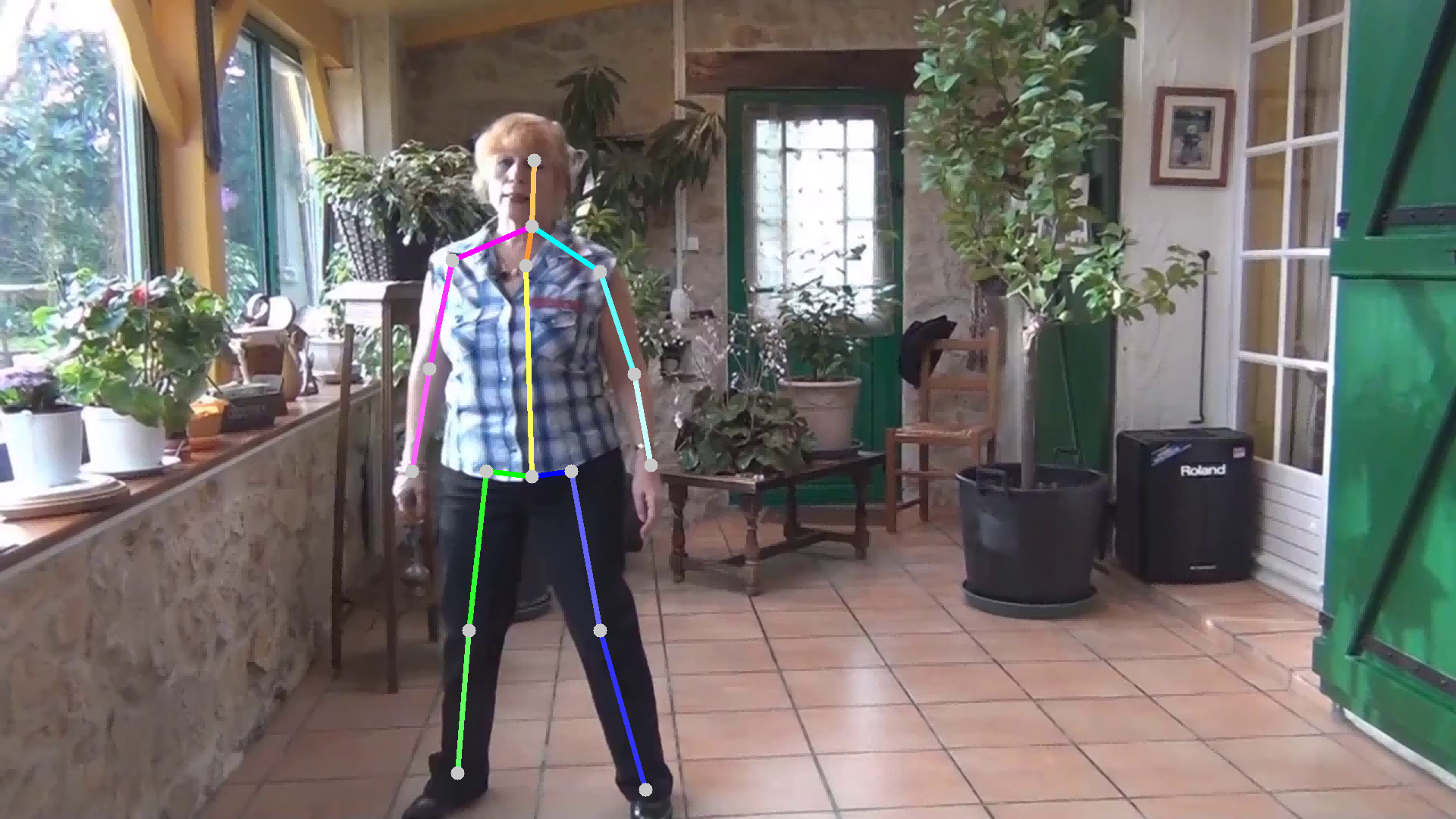}\vspace{4pt}
        \includegraphics[width=1\linewidth]{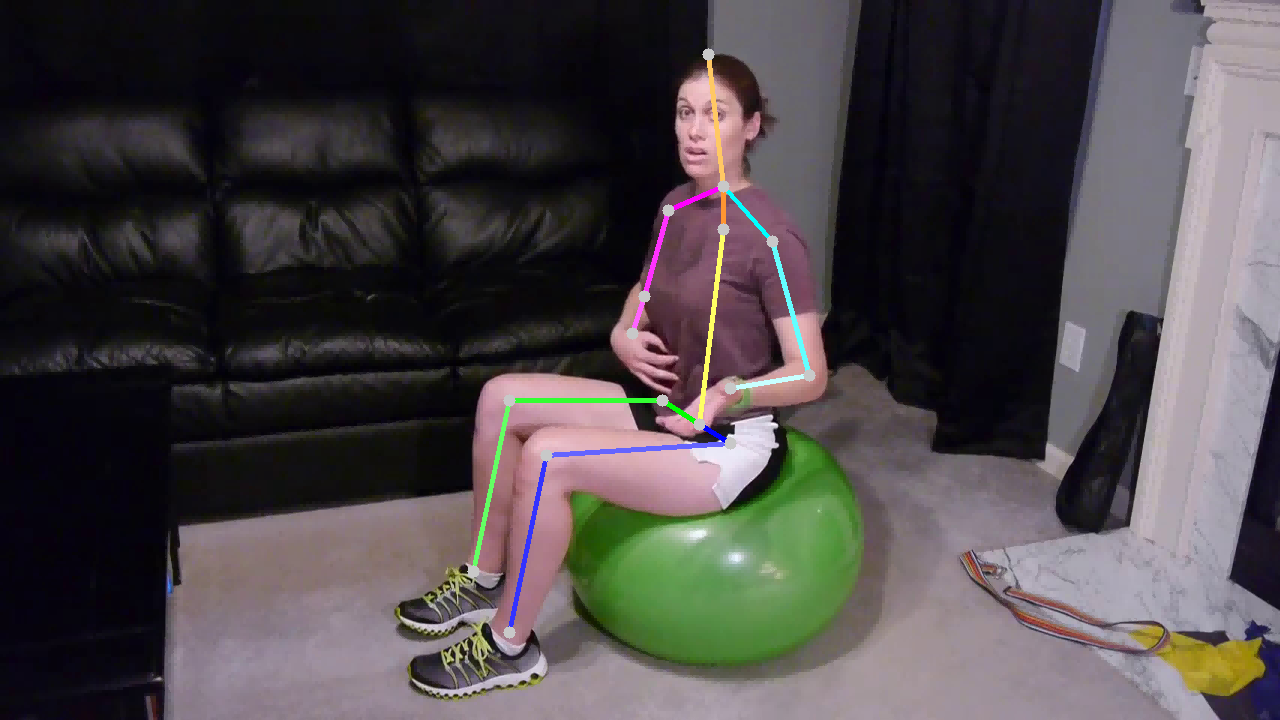}
    \end{minipage}}
    \caption{Visualization of human pose estimation results on the MPII validation dataset.}
\end{figure*}

\fi


\vspace{-15pt}
\begin{IEEEbiography}[{\includegraphics[width=1in,height=1.25in,clip,keepaspectratio]{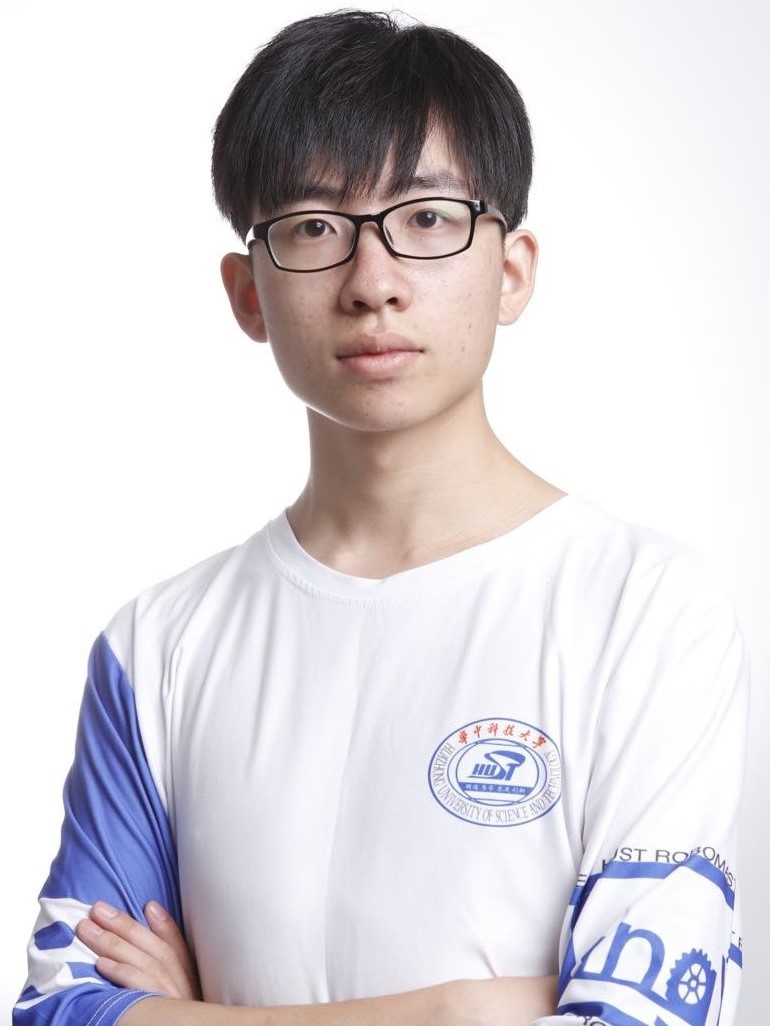}}]{Jiemin Fang}
    received the B.E. degree from School of Electronic Information and Communications, Huazhong University of Science and Technology, Wuhan, China, in 2018. He is currently a PhD candidate at the Institute of Artificial Intelligence and School of Electronic Information and Communications, Huazhong University of Science and Technology. His research interests include AutoML and efficient deep learning.
\end{IEEEbiography}

\vspace{-15pt}
\begin{IEEEbiography}[{\includegraphics[width=1in,height=1.25in,clip,keepaspectratio]{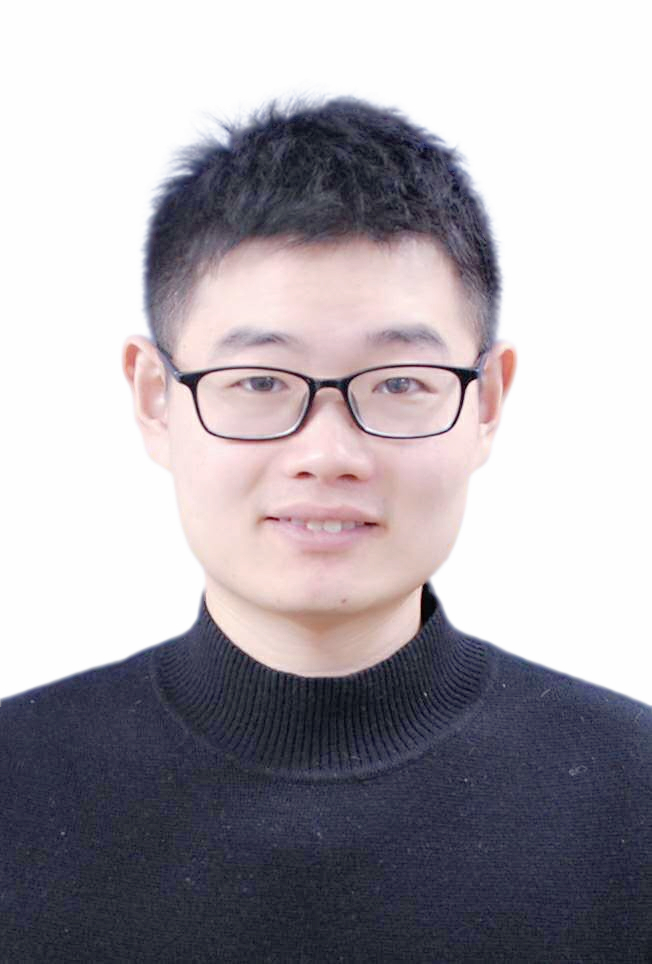}}]{Yuzhu Sun}
    received the B.E. degree from School of Electronic Information and Communications, Huazhong University of Science and Technology, Wuhan, China, in 2019. He is currently a master student at School of Electronic Information and Communications, Huazhong University of Science and Technology. His research interests include semantic segmentation and neural architecture search.
\end{IEEEbiography}

\vspace{-15pt}
\begin{IEEEbiography}[{\includegraphics[width=1in,height=1.25in,clip,keepaspectratio]{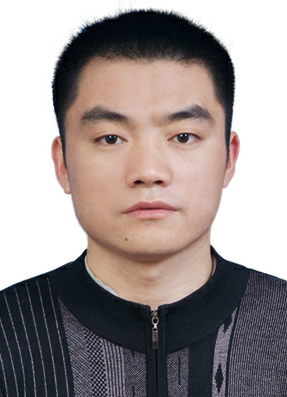}}]{Qian Zhang}
    received the B.E. and M.S. degrees from Central South University, Changsha, China, in 2008 and 2011, respectively, and the Ph.D. degree in pattern recognition and intelligent systems from the Institute of Automation, Chinese Academy of Sciences, Beijing, China, in 2014. His current research interests include computer vision and machine learning.
\end{IEEEbiography}

\vspace{-15pt}
\begin{IEEEbiography}[{\includegraphics[width=1in,height=1.25in,clip,keepaspectratio]{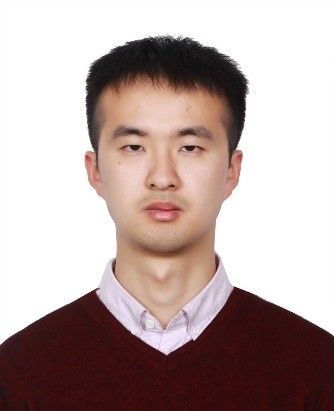}}]{Kangjian Peng}
    received the B.E. and M.S. degrees from Hangzhou Dianzi University, Hangzhou, China, in 2016 and 2019, respectively. He is currently a software engineer in Horizon Robotics. His research interests include neural architecture search and object detection.
\end{IEEEbiography}

\vspace{-15pt}
\begin{IEEEbiography}[{\includegraphics[width=1in,height=1.25in,clip,keepaspectratio]{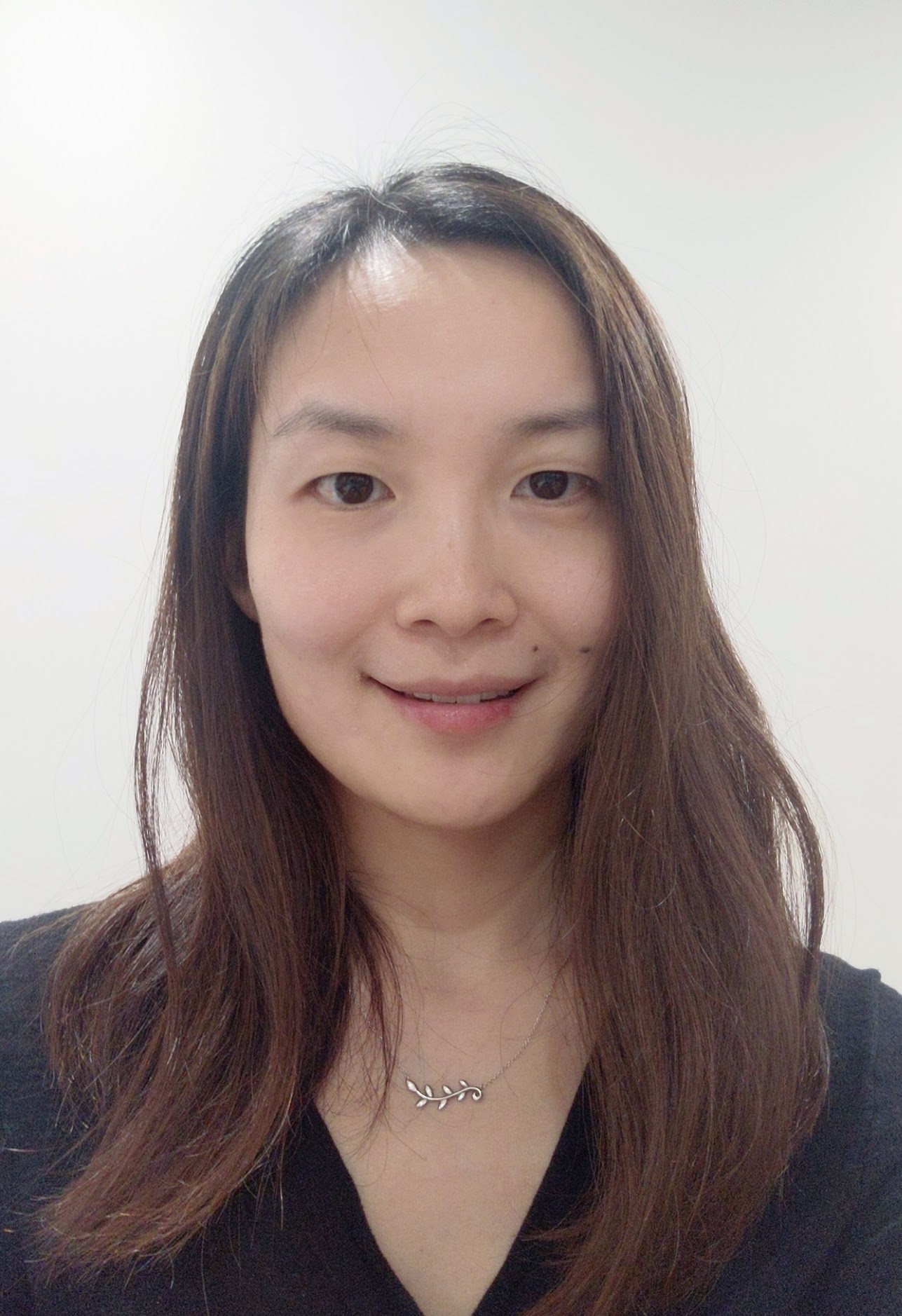}}]{Yuan Li}
    received her Bachelor's and Master's degrees in Computer Science from Tsinghua University and University of Southern California in 2005 and 2008 respectively. Since graduate school she has been working on computer vision. She is currently a senior engineering manager at Google Research. Her research interest includes human sensing and object understanding.
\end{IEEEbiography}

\vspace{-15pt}
\begin{IEEEbiography}[{\includegraphics[width=1in,height=1.25in,clip,keepaspectratio]{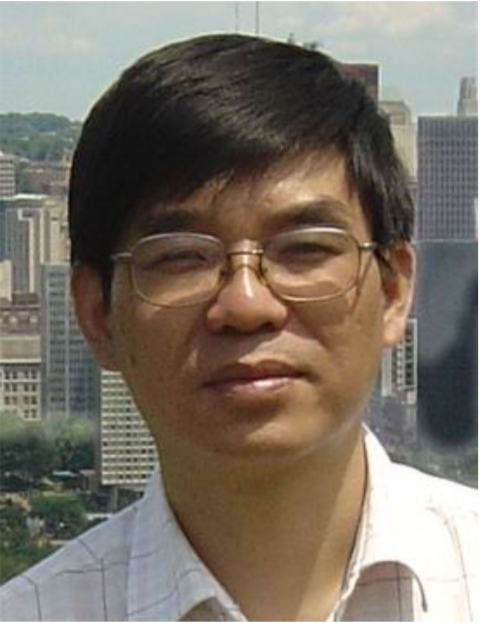}}]{Wenyu Liu (SM’15)}
    received the B.S. degree in Computer Science from Tsinghua University, Beijing, China, in 1986, and the M.S. and Ph.D. degrees, both in Electronics and Information Engineering, from Huazhong University of Science and Technology (HUST), Wuhan, China, in 1991 and 2001, respectively. He is now a professor and associate dean of the School of Electronic Information and Communications, HUST. His current research areas include computer vision, multimedia, and machine learning.    
\end{IEEEbiography}

\vspace{-15pt}
\begin{IEEEbiography}[{\includegraphics[width=1in,height=1.25in,clip,keepaspectratio]{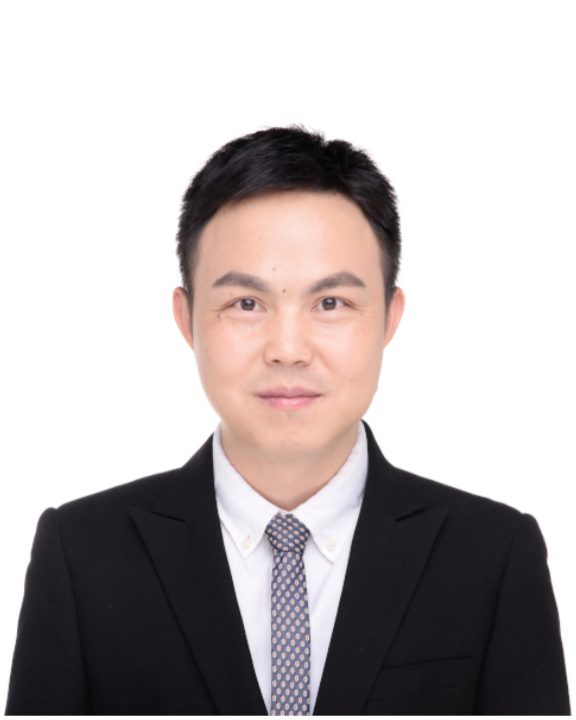}}]{Xinggang Wang}
    received the B.S. and Ph.D. degrees in Electronics and Information Engineering from Huazhong University of Science and Technology (HUST), Wuhan, China, in 2009 and 2014, respectively. He is currently an Associate Professor with the School of Electronic Information and Communications, HUST. His research interests include computer vision and machine learning.
\end{IEEEbiography}




\end{document}

%% file: math_commands.tex
\usepackage{amsmath,amsfonts,bm}

\def\eqref#1{equation~\ref{#1}}
\def\1{\bm{1}}

\def\va{{\bm{a}}}

\def\vo{{\bm{o}}}

\def\vv{{\bm{v}}}

\def\vx{{\bm{x}}}

\def\mO{{\bm{O}}}

\def\mW{{\bm{W}}}
\def\mX{{\bm{X}}}

\DeclareMathAlphabet{\mathsfit}{\encodingdefault}{\sfdefault}{m}{sl}
\SetMathAlphabet{\mathsfit}{bold}{\encodingdefault}{\sfdefault}{bx}{n}

\def\gN{{\mathcal{N}}}

\newcommand{\R}{\mathbb{R}}

\newcommand{\normlone}{L^1}